\documentclass[10pt,journal,compsoc]{IEEEtran}
\usepackage{graphicx}
\usepackage{adjustbox}
\usepackage{subcaption}
\usepackage{tikz}
\usepackage{tabularx}
\usepackage[T1]{fontenc}
\usepackage[utf8]{inputenc}
\usepackage{lmodern}
\usepackage{amsmath}
\usepackage{amssymb}
\usepackage{url}
\usepackage{color}
\usepackage{enumitem}
\usepackage{algcompatible}
\usepackage[boxed,ruled,titlenumbered,linesnumbered,figure]{algorithm2e}
\usepackage{rotating}
\usepackage{color}
\usepackage{tabularx}
\usepackage{bm}

\let\oldnl\nl
\newcommand{\nonl}{\renewcommand{\nl}{\let\nl\oldnl}}

\ifCLASSOPTIONcompsoc
  \usepackage[nocompress]{cite}
\else
  \usepackage{cite}
\fi

\begin{document}


\title{Relationship between Variants of One-Class Nearest Neighbours and Creating their Accurate Ensembles}

\author{Shehroz S. Khan,
        Amir Ahmad
        \thanks{Dr. Shehroz Khan is currently affiliated with the University of Toronto, Canada, shehroz.khan@utoronto.ca}
        \thanks{Dr. Amir Ahmad is affiliated with the College of Information Technology, United Arab Emirates University, Al Ain City, UAE, amirahmad@uaeu.ac.ae}}

%
 
\IEEEtitleabstractindextext{%
\begin{abstract}
In one-class classification problems, only the data for the target class is available, whereas the data for the non-target class may be completely absent. In this paper, we study one-class nearest neighbour (OCNN) classifiers and their different variants. We present a theoretical analysis to show the relationships among different variants of OCNN that may use different neighbours or thresholds to identify unseen examples of the non-target class. 
We also present a method based on inter-quartile range for optimising parameters used in OCNN in the absence of non-target data during training. Then, we propose two ensemble approaches based on random subspace and random projection methods to create accurate OCNN ensembles. We tested the proposed methods on $15$ benchmark and real world domain-specific datasets and show that random-projection ensembles of OCNN perform best.
\end{abstract}

}

\maketitle
\IEEEpeerreviewmaketitle

\IEEEraisesectionheading{\section{Introduction}}
One-Class Classification (OCC) \cite{khan2009survey} is a special case of  classification problem where the data of the positive (target) class is sufficiently available, whereas the negative class (non-target) data is absent during the training phase. 
Some examples for OCC are machine fault diagnosis, fraud detection, human fall detection, identifying a rare disease, etc.  \cite{Khan:KER:2014}. In most of these applications, it is easy to collect samples for the positive class or the data describing the normal behaviour of the domain under consideration. However, the collection of negative samples may be very difficult or result in high cost in dollars or put the health and safety of a person in danger \cite{Khan:KER:2014}. In other cases, the data in the negative class occur rarely. Therefore, even if some samples are collected for the negative class, the training data will be severely skewed and it is difficult to build generalisable classifiers using the  traditional supervised classification algorithms. Formally, Tax \cite{Tax2001}, defined a one-class classifier as:
\[f(z)= I(d(z) < \theta_d)\]
where $d(z)$ is a measure of distance (or probability) of an object $z$ to the target class (represented by a training set T), $\theta_d$ is a threshold on  $d(z)$, $I(.)$ is an indicator function and $f(z)$ represents a function which accepts new objects when the distance to the target class is smaller than $\theta_d$.

The research on OCC problems leads to several approaches to build classification models based on only the positive samples 
using support vector machines, nearest neighbours, decision trees, bayesian networks, neural networks and classifier ensemble approaches \cite{Khan:KER:2014}. In this paper, we choose a one-class nearest neighbour (OCNN) approach for detecting unseen samples of the negative class, when they are not present during the training phase. 
An OCNN method finds the high and low density regions based on the local neighbourhood of a test sample. Using a decision threshold, an OCNN accepts or rejects a test sample as a member of the target class. In its simplest form, an OCNN finds the first nearest neighbour of a test sample in the target class and then finds the first nearest neighbour of this neighbour in the target class \cite{Tax2001}. If the ratio of these distances is lower than a user-defined threshold, it is accepted as a member of the target class \cite{tax1999support}. This method is simple and effective in finding instances of the unseen negative class. The OCNN does not require training the classifier, has few parameters and can easily be used with different similarity measures. These properties make OCNN a useful method for OCC problems.
In an OCNN, the number of nearest neighbours and the value of decision threshold can be optimised but the sensitivity of the classifier w.r.t. noise in the positive class may change. There exists several variants of OCNN in the literature (see Section \ref{sec:related_work}); however, these methods do not highlight relationship between different OCNN methods and do not clearly explain the reasons for a particular variant to work better than the other. 

Classifier ensemble approaches have been suggested for different OCC  methods to improve their performance \cite{Khan:KER:2014}. 
Previous research shows that classifier ensembles consisting of accurate and diverse classifiers perform better than a single classifier \cite{Kuncheva}. 
 The supervised counterpart of OCNN, i.e., K-nearest neighbour (KNN) is robust to  variations in the data set; therefore, ensemble techniques based on data sampling (such as Bagging and Boosting) have not been successful to create diverse KNN classifiers \cite{domeniconi2004nearest}. However, KNN classifiers are sensitive to input features sampling; therefore, diverse and accurate KNN based ensembles can be created by using different feature spaces \cite{RPnsemblenearestneigh}. We propose that similar ensemble approaches can be useful for OCNN ensembles.
In this paper, we present a holistic view on the general problem of OCNN by discussing their different variants and creating their accurate ensembles; the main contributions of the paper are:
\begin{itemize}[leftmargin=*] 
  \item We theoretically show the relationship between various OCNN approaches and discuss the impact of choosing nearest neighbours on the decision threshold. 
  \item We present a cross-validation method and a modified thresholding algorithm that utilise the outliers from the target class to optimise the parameters of the OCNN.
  \item We present two different types of OCNN ensemble methods by using random subspace and random projection methods to study their performance when the feature space is either changed or transformed. To the best of our knowledge, the suitability of random projection approach is investigated for the first time for OCNN. 
\end{itemize}

Our results on several benchmark and domain-specific real world datasets show superior performance of the OCNN with random projection ensembles in comparison to single OCNN.

The rest of the paper is structured as follows. In Section \ref{sec:related_work}, we present literature review on different variants of OCNN and their ensembles. Section \ref{sec:occjknn} introduces two variants of OCNN that uses different numbers of nearest neighbours and decision threshold, followed by a theoretical analysis about their relationship with other variants of OCNN. In Section \ref{sec:procedure}, we introduce a cross-validation method and a modified thresholding algorithm to optimise parameters for the OCNN classifiers using only the data from the target class. In Section \ref{sec:ensemble},  we present a discussion on various ensemble approaches that are used with the OCNN classifiers. Experimental methods and results are shown on various datasets in Section \ref{sec:results} and \ref{sec:experiment}. Conclusions and future work are summarised in Section \ref{sec:conclusions}.

\section{Related Work}
The research on OCC leads to the development of various models that mostly differ in either learning with target examples only; learning with target examples and some amount of poorly sampled outlier examples or artificially generated outliers; learning with labeled target and unlabeled data; methodology used and application domains applied \cite{khan2009survey}. In this paper, we restrict our literature review to OCNN approaches and the methods that involve their ensemble. 

\label{sec:related_work}
Tax \cite{Tax2001} presents an OCNN method called nearest neighbour description (\textit{NN-d}). 
The general idea is to find the distance of a test object to its nearest neighbour in the target class, and find the nearest neighbour of this neighbour, and if the distance is greater than a threshold then it is identified as an outlier. Tax and Duin \cite{tax2000data} propose a nearest neighbour method capable of finding data boundaries when the sample size is very low. The boundary thus constructed can be used to detect the targets and outliers. However, this method has the disadvantage that it relies on the individual positions of the objects in the target set. Their method seems to be useful in situations where the data is distributed in subspaces. They test the technique on both real and artificial data and find it to be useful when very small amounts of training data exist (fewer than 5 samples per feature).
Haro-Garc{\i}a et al. \cite{haro2008one} use \textit{NN-d} along with other one-class classifiers for identifying plant/pathogen sequences. They find the suitability of above methods owing to the fact that genomic sequences of plant are easy to obtain in comparison to pathogens. Their basic idea is to optimise the threshold between the distances of a test object to its nearest neighbour and distance of this neighbour to its nearest neighbour. They order these distances and find the best threshold based on the percentile of the sorted sequence. Angiulli \cite{angiulli2012prototype} present a prototype based domain description algorithm (\textit{PDD}) for OCC tasks, which is equivalent to \textit{NN-d} under the infinity Minkowski metric and generalises statistical tests for outlier detection. Angiulli introduces the concept of \textit{PDD} consistent subset prototypes, which is a subset of the original prototype set. The paper presents two algorithms for computing \textit{PDD} consistent subset: one that guarantees logarithmic approximation factor and the other faster version to manage very large datasets. The faster version does not consider all the pairwise object distances but it is shown to return almost identical subset size at lower computational cost.  
Krawczyk et al. \cite{Krawczyk2014A} use prototype reduction techniques using evolutionary approaches that follow either selection or generation of examples from the training data and perform \textit{NN-d} approach. They report no significant accuracy losses using their prototype reduction technique on several datasets from UCI repository. Cabral et al. \cite{cabral2007novel} propose an OCNN method that uses the concept of structural risk minimisation. 
They remove redundant samples from the training set, thereby obtaining a compact representation aiming at improving generalisation performance of the classifier. They achieve considerable reduction in the number of stored prototypes and improved performance than the \textit{NN-d} classifiers.

Cabral et al. \cite{cabral2009combining} extend their work \cite{cabral2007novel} and present another approach where they not only consider one but $K$-nearest neighbours of a test object and find a nearest neighbour of each of these $K$ neighbours. A decision is arrived based on majority voting. Their results show that $K$ nearest neighbour version of their classifier outperforms the \textit{NN-d}. However, it is not shown how the optimal value of $K$ nearest neighbour is selected for better classification results. Munroe and Madden \cite{munroe2005multi} extend the idea of \textit{NN-d} to tackle the recognition of vehicles using a set of features extracted from their frontal view and presented high accuracy of classification. They compute the average of the $K$ nearest neighbours of the first nearest neighbour to arrive at a decision. However, the computation of best fitting value of $K$ is not illustrated clearly in their work. Halvani et al. \cite{Halvani2013authorship} use the similar idea for authorship verification task. They generate different types of features for a given document. For every feature type, the distance is calculated for its nearest neighbour and average distance of its $K$ nearest neighbours, and a decision is taken for a given threshold. The decisions from all the feature types are combined using majority voting rule to take a final decision. Huang et al. \cite{huang2013prediction} present an OCNN method for the identification of protein carbamylated that play important role in a number of biological conditions. They compute threshold from the positive class based on a fixed false negative ratio, compute distance of a test sample with its $K$ neighbours and if it is below a threshold, it is accepted as a member of the positive class. 
Ges{\`u} et al. \cite{Gesu2008Knowledge} present an OCNN algorithm and tested it on synthetic data that simulate microarray data for the identification of nucleosomes and linker regions across DNA. They present a decision rule which states that if there are at least $K$ data objects in the positive class that are dissimilar from the test object at most $\phi$, then it is classified as a member of the positive class. They propose to calculate optimal values for $K$ and $\phi$ using the ROC curve. Their results show good recognition rate on synthetic data for nucleosome and linker regions across DNA. However, the paper does not explain the construction of validation set and parameters optimisation in detail. 
Khan \cite{khan2010kernels} presents a variant of the \textit{NN-d} method that finds the distances of $J$ nearest neighbours of  the test sample in the target class and compute the $K$ nearest neighbours of each of the $J$ neighbours in the target class. The distance ratio is computed for each of the $J$ neighbours and a majority voting rule is used to take a final decision. They use Gaussian and polynomial kernel as a distance metrics and show that this method can perform better than the traditional \textit{NN-d} on chemical spectral data.

A single one-class classifier might not capture all the characteristics of the data. Therefore, an ensemble of one-class classifiers is a viable solution to improve the performance of base one-class classifiers, which may differ in complexity or in the underlying training algorithm used to construct them \cite{Khan:KER:2014}.
P\c{e}kalska et al. \cite{pekalska_combining_2004} use the proximity of target object to its class as a `dissimilarity representation' (DR) and show that the discriminative properties of various DRs can be enhanced by combining them properly. They use three types of one-class classifier, namely \textit{NN-d}, Generalised Mean Class and Linear Programming Dissimilarity Data Description. They make two types of ensembles: (i) combine different DR from individual one-class classifiers into one representation after proper scaling using fixed rules, for e.g. average, product and train single one-class classifier based on this information, (ii) combine different DR of training objects over several base classifiers using majority voting rule. Their results show that both methods perform significantly better than the one-class classifiers trained with a single representation. 
Segu{\'\i} et al. \cite{segui2013bagged} present an ensemble method based on non-parametric weighted bagging strategy for OCC. Their method estimates a probability density based on a forest structure of the data instead of assuming uniform data distribution and constructs bootstrap samples according to the data density ranking, where higher density imply greater likelihood of being selected for a bootstrap of the ensemble. They use three types of one-class classifiers: \textit{NN-d}, One-Class Support Vector Machine (OSVM) and Minimum Spanning Tree Class Descriptor (MST) \cite{juszczak_minimum_2009} and show experimentally that the ensemble bagging and non-parametric weighted bagging obtain better rankings than the base one-class classifier methods and show a statistically significant improvement for MST and \textit{NN-d}. Menahem et al. \cite{Menahem2013Combining} propose an ensemble meta-learning algorithm which learns a combining function upon aggregates of the ensemble-members prediction. This algorithm depends on the classification properties of the ensemble-members and not on the fix-rule scheme. They use four one-class classifiers; two of them are based on OCNN, the rest are based on density estimation and OSVM. They combine these classifiers by four ensembles rules (majority voting, mean voting, max rule and product rule) and show that their proposed ensemble algorithm outperform traditional ensemble schemes. Ges{\`u} and Bosco \cite{d._gesu_combining_2007} present an ensemble method for combining fuzzy OCNN classifiers. Their classifier combining method is based on a genetic algorithm optimisation procedure by using different similarity measures. They test their method on two categorical datasets and show that whenever the optimal parameters are found, fuzzy combination of one-class classifiers may improve the overall recognition rate. Nanni \cite{nanni_experimental_2006} studies combining several one-class classifiers using the random subspace method for the problem of online signature verification. 
Nanni uses several one-class classifiers: Gaussian model description, Mixture of Gaussian Descriptions, \textit{NN-d}, PCA Description (PCAD), Linear Programming Description (LPD), Support Vector Data Description and Parzen Window Classifier. It is shown that fusion of various classifiers can reduce the error and the best fusion method is the combination of LPD and PCAD.  


\begin{figure}
\centering
\begin{tikzpicture}[sibling distance=10em,
  every node/.style = {shape=rectangle, rounded corners,
    draw, align=center,
    top color=white, bottom color=blue!20}]]
  \node {OCNN}
    child { node {Single Classifiers} 
      child {node {\cite{Tax2001},\cite{haro2008one},\cite{angiulli2012prototype},\\ \cite{Krawczyk2014A},\cite{cabral2007novel},\cite{munroe2005multi},\\ \cite{Halvani2013authorship},\cite{huang2013prediction}, \cite{Gesu2008Knowledge},\\ \cite{cabral2009combining},\cite{khan2010kernels}}}
      }
    child { node {Ensemble Classifiers}
      child { node {\cite{pekalska_combining_2004},\cite{segui2013bagged},\cite{juszczak_minimum_2009},\\\cite{Menahem2013Combining},\cite{d._gesu_combining_2007},\cite{nanni_experimental_2006}} }
      };
\end{tikzpicture}
  \caption{Summary of Literature Review.}
  \label{fig:litrev}
\end{figure}
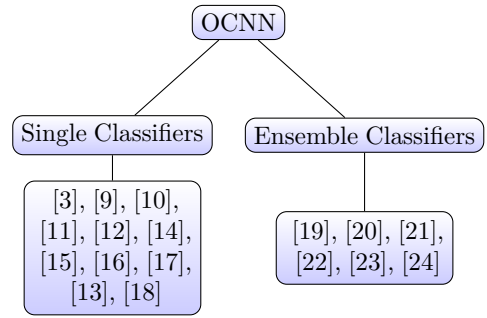

A graphical summary of literature review is shown in Figure \ref{fig:litrev}. 
The literature review suggests that different researchers use one or more nearest neighbours in OCNN to decide if a test instance is a member of the negative class or not. However, the intuition behind \textit{which} method will work better under \textit{what} conditions is not clear. There is not much work done on the optimisation of the decision threshold of an OCNN, which is either set arbitrarily or equal to $1$. A major challenge in most of the OCC methods is the unavailability of data from the negative class; therefore, optimisation of parameters is very difficult \cite{LiuModular2016}. Hence, optimising different nearest neighbours and the decision threshold is very difficult in the OCNN approach. We also observed that the ensembles of OCNN (such as random subspace) and fusion with other one-class classifiers can perform better than the base OCNN. Below, we present a detailed analysis of different variants of OCNN and perform a theoretical analysis on their relationship.

\section{One-Class Nearest Neighbour Classifiers}
\label{sec:occjknn}
Based on the literature review in Section \ref{sec:related_work}, various OCNN methods can be categorised into the following four types based on the number of nearest neighbours they use to compute the decision threshold: 

\begin{enumerate}[label=(\roman*)]
  \item Find the first nearest neighbour of the test sample in the target class, and the first nearest neighbour of the first neighbour ($11NN$) \cite{Tax2001,haro2008one,angiulli2012prototype,Krawczyk2014A,cabral2007novel}.
  \item Find the first nearest neighbour of the test sample in the target class, and the $K$ nearest neighbours of the first neighbours ($1KNN$) \cite{munroe2005multi,Halvani2013authorship,huang2013prediction}.
  \item Find the $J$ nearest neighbours of the test sample in the target class, and the first nearest neighbours of the first $J$ neighbours ($J1NN$) \cite{cabral2009combining}.
  \item Find the $J$ nearest neighbours of the test sample in the target class, and the $K$ nearest neighbours of the first $J$ neighbours ($JKNN$) \cite{khan2010kernels}.  
\end{enumerate}

\begin{figure}[!ht]
  \begin{subfigure}[b]{0.24\textwidth}
    \centering
    \includegraphics[width=3cm, height=4cm]{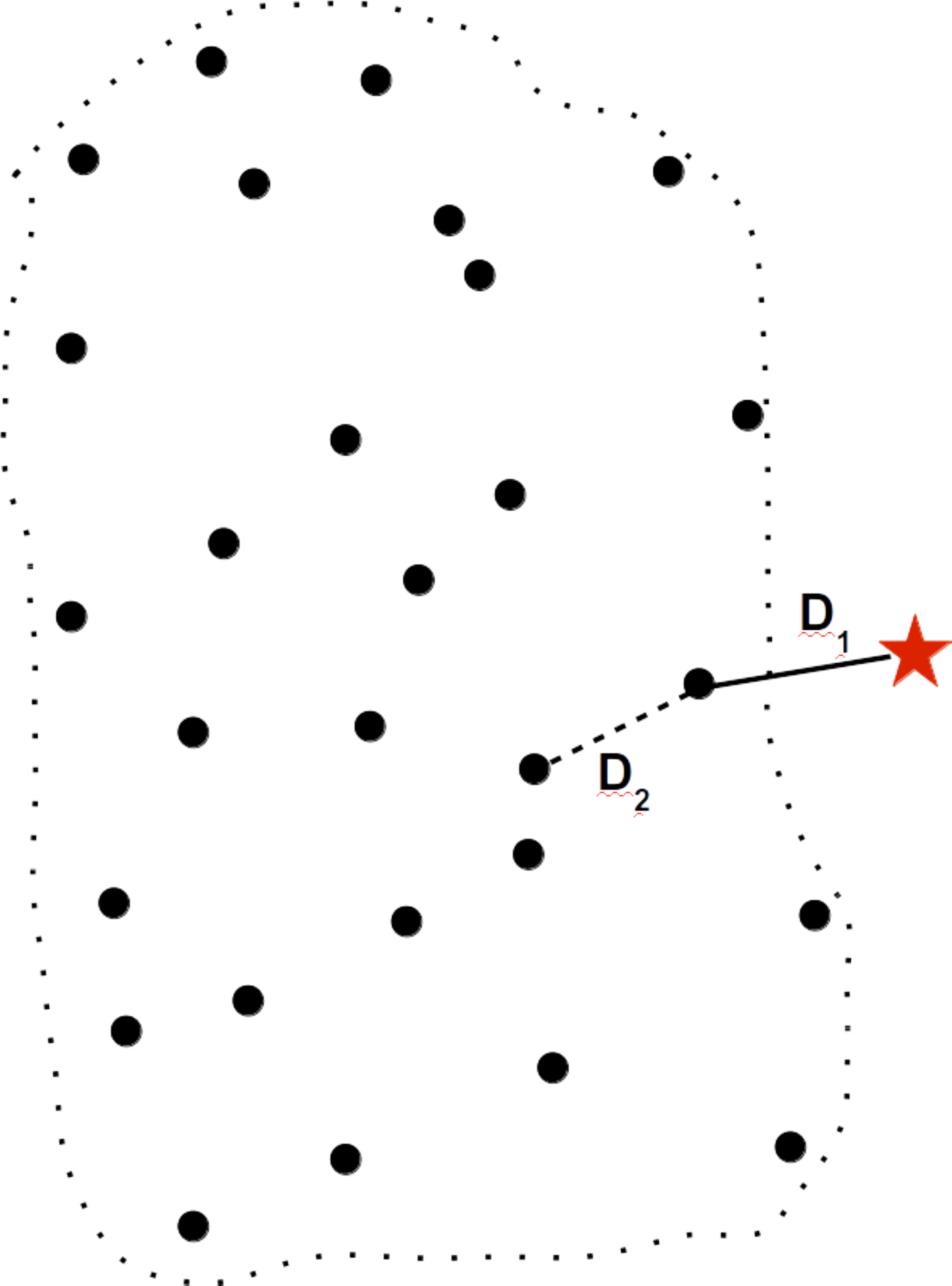}
    \caption{$11NN$ Nearest Neighbour}
    \label{fig:11nn}
  \end{subfigure}%
  ~
  \begin{subfigure}[b]{0.24\textwidth}
    \centering
    \includegraphics[width=3cm, height=4cm]{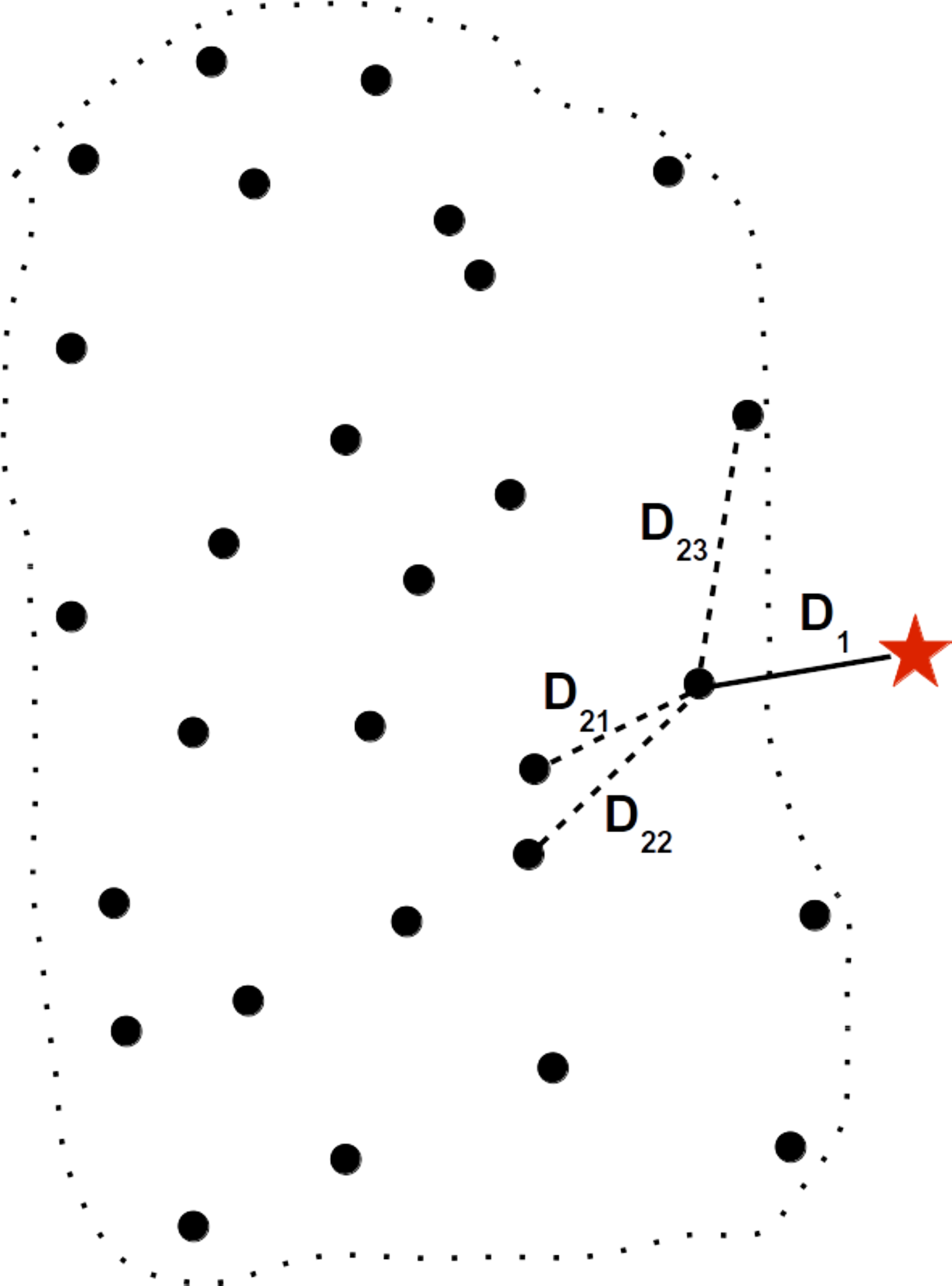}
    \caption{$1KNN$ Nearest Neighbour}
    \label{fig:1knn}
  \end{subfigure}%
  
  \begin{subfigure}[b]{0.24\textwidth}
    \centering
    \includegraphics[width=3cm, height=4cm]{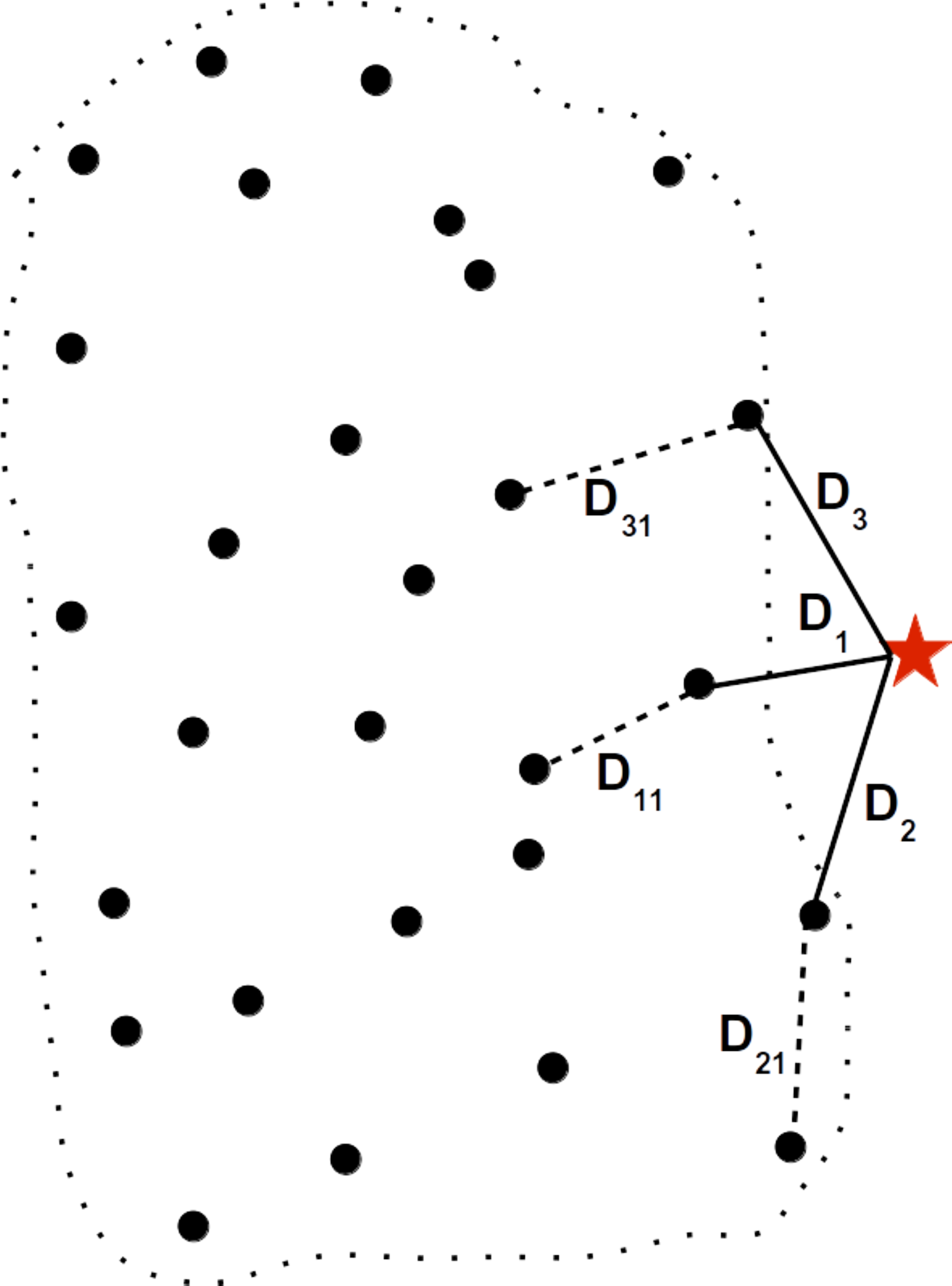}
    \caption{$J1NN$ Nearest Neighbour}
    \label{fig:j1nn}
  \end{subfigure}%
  ~
  \begin{subfigure}[b]{0.24\textwidth}
    \centering
    \includegraphics[width=3cm, height=4cm]{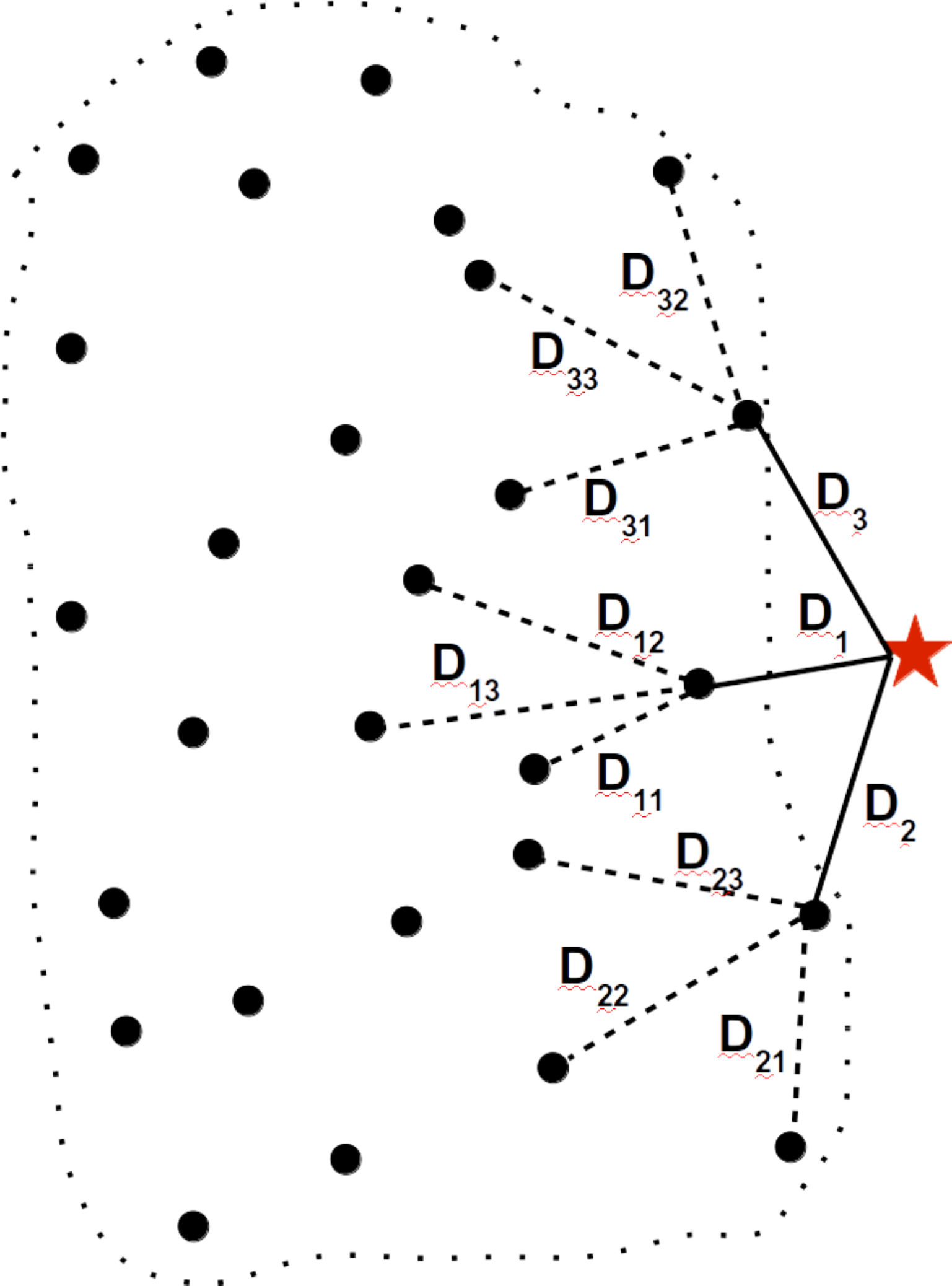}
    \caption{$JKNN$ Nearest Neighbour}
    \label{fig:jknn}
  \end{subfigure}%
  \caption{Different variants of OCNN methods.}
  \label{fig:ocnn}
\end{figure}

These different methods of OCNN can differ in the number of nearest neighbours, distance metric, method of combining distances and the choice of the value of the decision threshold to arrive at a final decision. Figure \ref{fig:ocnn} shows a graphical representation of these methods using artificial data in $2$ dimension. The dark circles ($\bullet$) show the instances of the target class. The red star ({\color{red}$\star$}) shows a test sample. The solid line shows the distance of either $1$ or $J$ nearest neighbours of the test sample in the target class and the dotted lines show the distance between the $1$ or $K$ nearest neighbours of those $1$ or $J$ neighbours. In a $JKNN$ classifier, if $J$ and/or $K$ is set to $1$, then it condenses to $11NN$, $1KNN$ or $J1NN$.

\subsection{One-Class JK Nearest Neighbour} 
\label{sec:jknn}

We now present a general OCNN ($JKNN$) method for detecting non-target samples. In this method, we 
\begin{enumerate}[leftmargin=*]
  \item Find the $J$ nearest neighbour ($NN_j^{tr}(z)$) of the test sample ($z$) in the target class and find their average, $\bar{D}_J$.
  \item Find the $K$ nearest neighbours of these $J$ neighbours ($NN_k^{tr}(NN_j^{tr}(z))$) and find their average, $\bar{D}_K$.
\end{enumerate}

For a given decision threshold, $\theta$, if $\frac{\bar{D}_J}{\bar{D}_K}<\theta$, then the test data object is considered as a member of the target class or else rejected as a member of the negative class. 
Algorithm \ref{algo:jknn} shows the steps to classify a test data object as a member of target class or not. This algorithm is different from the work of Khan \cite{khan2010kernels} that use a similar $JKNN$ variant in two ways. Firstly, they use kernel (Gaussian and polynomial) a distance metric, whereas we use euclidean as a distance metric. Secondly, they combine the decision using majority voting of the distances of different $K$ neighbours of each $J$ neighbours, whereas we take the average of the distances of the $J$ and $K$ nearest neighbours.

  \begin{algorithm}
    \DontPrintSemicolon
    \KwIn{$N$ number of $T$ target data objects with dimension $d$, Test sample $z$, \textit{J} and \textit{K} neighbours, $\theta$ threshold}
    \KwOut{Decision:\textit{reject} or \textit{accept} the test sample as member of target class}
    Compute the distance of a test sample, $z$ to its \textit{J} nearest neighbours and find their average,
    $\bar{D}_J = \frac{1}{J}\sum_{j=1}^J \Vert z- NN_j^{tr}(z)\Vert$\\
    Compute the distance of each $j^{th}$ nearest neighbour to its \textit{K} nearest neighbours, and find their average
    $\bar{D}_K = \frac{1}{J*K} \sum_{j=1}^J \sum_{k=1}^K \Vert NN_j^{tr}(z)- NN_k^{tr}(NN_j^{tr}(z))\Vert$\\
    \eIf{$\frac{\bar{D}_J}{\bar{D}_K}<\theta$}{
      \Return{\textit{accept}}
      }{
      \Return{\textit{{reject}}
      }
      }
    \caption{One-Class \textit{J-K} Nearest Neighbour}
  \label{algo:jknn}
   \end{algorithm}  

%

\subsection{Relationship Among Different OCNN approaches}
\label{sec:relation}
In an OCNN, either the decision threshold ($\theta$) is kept as $1$ or chosen arbitrarily.
In this section, we will show that varying decision threshold ($\theta$) with $11NN$ is similar to other OCNN methods discussed in Section 3.

In Figure \ref{fig:11nn}, for $11NN$ with $\theta=1$,   if $D_1$ is more than  $D_2$ the test sample is classified as an outlier. In this case even if $D_1$ is slightly more than  $D_2$ the test sample will be assigned to negative class.  Intuitively, an outlier data point should be at a much greater distance to its nearest neighbour ($D_1$) than the distance between this nearest neighbour and its nearest neighbour ($D_2$).  Mathematically this can be represented as 
\begin{equation}
  \label{eq:1knntheta}
 D_1 > \theta D_2
  \end{equation}
where $\theta$  > 1.                                 
In this case, many data points that are outliers in the first case ($D_1 > D_2$), will not be rejected. The $D_1 > \theta D_2$ rule should reject less outliers than  $D_1 > D_2$ rule because an outlier's dissimilarity with its nearest data point in the target class should be more than the dissimilarity between that nearest data point to its nearest neighbour (we know that these two data points belong to the target class).  However, finding the optimal value of $\theta$ is problematic as it depends on the properties of a dataset. The optimal value of $\theta$ may be different for different datasets and depends on the performance measure used. We will now discuss the relation between other variants of OCNN:

\begin{enumerate}[leftmargin=*]
  \item \textbf{$1KNN$ approach} -- The $1KNN$ approach can produce similar results as $11NN$ ($\theta$  > 1). From Figure \ref{fig:1knn}, for a point of negative class, we can write the distances as
  \begin{equation}
  \label{eq:1knn}
  D_1 > ((D_{21} +D_{22}....+D_{2i},......D_{2K})/K)
  \end{equation}
  
where $D_{2i}$ $(K\geq i \geq 1)$ is the distance between the first nearest neighbour of the test sample and the $i^{th}$ nearest neighbour of this nearest neighbour. As that distance between the first nearest neighbour and its  $i^{th}$ nearest neighbour is greater than or equal to the distance between it and  $(i-1)^{th}$ nearest neighbour, 
 
$\therefore D_{2i}\geq D_{2(i-1)} \implies D_{2i} \geq D_{21}$ 

Expanding the terms, we get  
  \begin{equation}
    \begin{split}
    \label{eq:1knnA}
    (D_{21} +D_{22}....+D_{23},......D_{2K})/K  &\geq\\  (D_{21} +D_{21}....+D_{21}......D_{21})/K \\
    \implies(D_{21} +D_{22}....+D_{23},......D_{2K})/K &\geq  KD_{21}/K \\
    \implies(D_{21} +D_{22}....+D_{23},......D_{2K})/K &\geq  D_{21} \\
    \implies(D_{21} +D_{22}....+D_{23},......D_{2K})/K &= \alpha D_{21},\\  \quad \text{for} \quad \alpha \ge 1
    \end{split}                             
  \end{equation} 
                             
Using Equations \ref{eq:1knn} and \ref{eq:1knnA}, we get
  \begin{equation}
  \label{eq:1knnB}
    D_1 > \alpha D_{21}   
  \end{equation}

It is highly unlikely that all the ($K-1$) nearest neighbours of the first nearest neighbour of the test sample have the same distances as its first nearest neighbour. Hence, $\alpha > 1$ for most of the cases. 
Equations \ref{eq:1knntheta}  and \ref{eq:1knnB} are the same, this suggests that $1KNN$ is the same as the intuition based calculation for $11NN$ with $\theta$ > 1.

 \item \textbf{$J1NN$ approach -} As shown in Fig. \ref{fig:j1nn}, $J1NN$ computes the average distance of $J$ nearest neighbours of a test sample, this value is compared with the average distance of one nearest neighbour of these $J$ nearest neighbours to get the final decision. The calculation for assigning a class to a data point will be opposite to the $1KNN$ approach; hence, the condition to assign a negative class to a data point will be 
\begin{equation}
  \label{eq:j1nnB}
   \alpha D_1 >  D_{21}   
  \end{equation}

 Eq. \ref{eq:j1nnB} with Eq. \ref{eq:1knntheta} suggest that $1JNN$ approach is similar to $\theta$ $\leq$ 1. 

  \item \textbf{$JKNN$ approach} --  In $JKNN$ approach, the average distance  of $J$ nearest neighbours  of a test sample is compared with the average distance of the $K$ nearest neighbours of each of these $J$ nearest neighbours. For a data point to be a negative class (see Figure \ref{fig:jknn})
\begin{equation}
  \label{eq:jknn}
   \bar{D}_J > \bar{D}_K
  \end{equation}

  Using the same argument as discussed for $1KNN$, we can write Equation \ref{eq:jknn} as,
\begin{equation}
  \label{eq:jknntheta}
   \beta D_{1} > \gamma D_{21} 
  \end{equation}
where $\beta$ and $\gamma \geq 1$. 
More number of neighbors ($J$ and $K$) increase the average values of the distances ($\beta D_{1}$ and $\gamma D{21}$). As $D_1$ and $D_{21}$ is constant for a given case, $\beta$ and $\gamma$ increases with $J$ and $K$ respectively. Equation \ref{eq:jknntheta} can be re-written as:

\begin{equation}
  D_{1}  > (\gamma/\beta) D_{21} 
  \end{equation}

For most of the cases, $\beta$ and $\gamma$ are more than $1$; therefore, it is likely, they will cancel out each other's effect for large values of $J$ and $K$. Hence, for large values of $J$ and $K$, they will generate a condition, $D_{1}  > D_{21}$; that is similar to $11NN$ approach with $\theta = 1$.
For smaller values of $J$, the value of $\beta$ will be near to $1$. Therefore, for small $J$ and large $K$, ($\gamma/\beta$) will be greater than $1$.   
\end{enumerate}

The above arguments imply that $1KNN$ or $JKNN$ with small $J$ and $\theta$ = 1 will be similar to $11NN$ with $\theta > 1$. Thus, they are more likely to reject less outliers than $11NN$ with $\theta = 1$. Whereas, $J1NN$ or $JKNN$ with a large value of $J$ may accept more non-targets as members of the target class;  similar to $11NN$ with $\theta < 1$. 
The methods $1KNN$, $J1NN$, and $JKNN$ will take longer time than $11NN$ due to the computation of additional $J$ or $K$ nearest neighbours in comparison to the $11NN$ method (more details in the next Section).

Finding the optimised values of $J$ and $K$ for $JKNN$ with $\theta$ = 1 or the optimised value of $\theta$ for $11NN$ is an important step in designing an efficient OCNN classifier. 
The parameter optimisation techniques are discussed in Section \ref{sec:procedure}.

\section{Parameter Optimisation for OCNN}
\label{sec:procedure}
Optimising parameters in one-class classifiers is difficult because only the data from the positive class is available during training.
In this section, we will discuss a method by which noisy samples within the positive class can be used to optimise the parameters of OCNN classifiers.
We note that the different variants of OCNN discussed in Section \ref{sec:occjknn} (shown in Figure \ref{fig:ocnn}) can suffer from either or both of the following problems:

\begin{enumerate}[leftmargin=*]
  \item False Positives -- The real world target data may contain noisy observations due to non-calibrated data collection devices, human errors in labeling or inadvertent artifacts \cite{khan2014xfactor}. The above discussed OCNN approaches ignore this fact and the presence of such deviant observations in the target class can make these classifiers sensitive to noise. This case can arise when most of the target data objects are dense toward their centre but some data objects are far from it. This behaviour can lead to outliers being accepted as members of the target class.
  \item False Negatives -- If noisy observations are removed from the dataset without properly adjusting the decision threshold, OCNN can reject a target test sample as an outlier.
\end{enumerate}

Yin et al. \cite{DBLP:journals/tkde/YinYP08} mention that in the OCC setting, due to the scarcity of negative data, it is a challenging problem to design a detection system that can reduce both the false positives and false negatives. In general, OCC methods are very sensitive to the choice of parameters \cite{Mack2014Can}. In the case of OCNN, choosing a particular value of $J$ and $K$ nearest neighbours or moving the decision threshold ($\theta$) can affect both the false positive and false negative rates. However, in a given OCC problem, a classifier's parameters cannot be optimised directly because there is no training set and validation set available for the negative class. Therefore, choosing the optimal parameters for a given OCC problem in the absence of non-target class is very challenging \cite{liu2016modular}. 
One approach to handle such scenario is to generate artificial data for the unseen negative class \cite{hempstalk_one-class_2008,Khan:2012:TDU,desir2013one} by assuming some distribution for the instances of the negative class. However, these approaches are prone to over-fitting. Other possibility is to build models for the unseen negative class based on the parameters of the positive class \cite{khan2014xfactor,quinn}. However, such approaches depend on the parameters estimated from the positive class. A true distribution for positive class is hard to estimate due to limited availability of data and the assumptions on the dataset. We take an alternative view by attempting to find classification boundary from the existing training data for the target class. We now present a method for optimising the parameters of an OCNN in the absence of training samples for the negative class by:
\begin{enumerate}[leftmargin=*]
  \item Removing the noisy observations from the target class using a method based on Inter-Quartile Range. 
  \item Using the rejected noise data as a proxy for the unseen negative class and performing a new cross-validation method to optimise the best $J$ and $K$ neighbours or the decision threshold, $\theta$.
\end{enumerate}


\subsection{Removing noise from the target class} 
\label{sec:iqr}
\begin{figure}
  \centering
  \includegraphics[width=9cm,height=9cm]{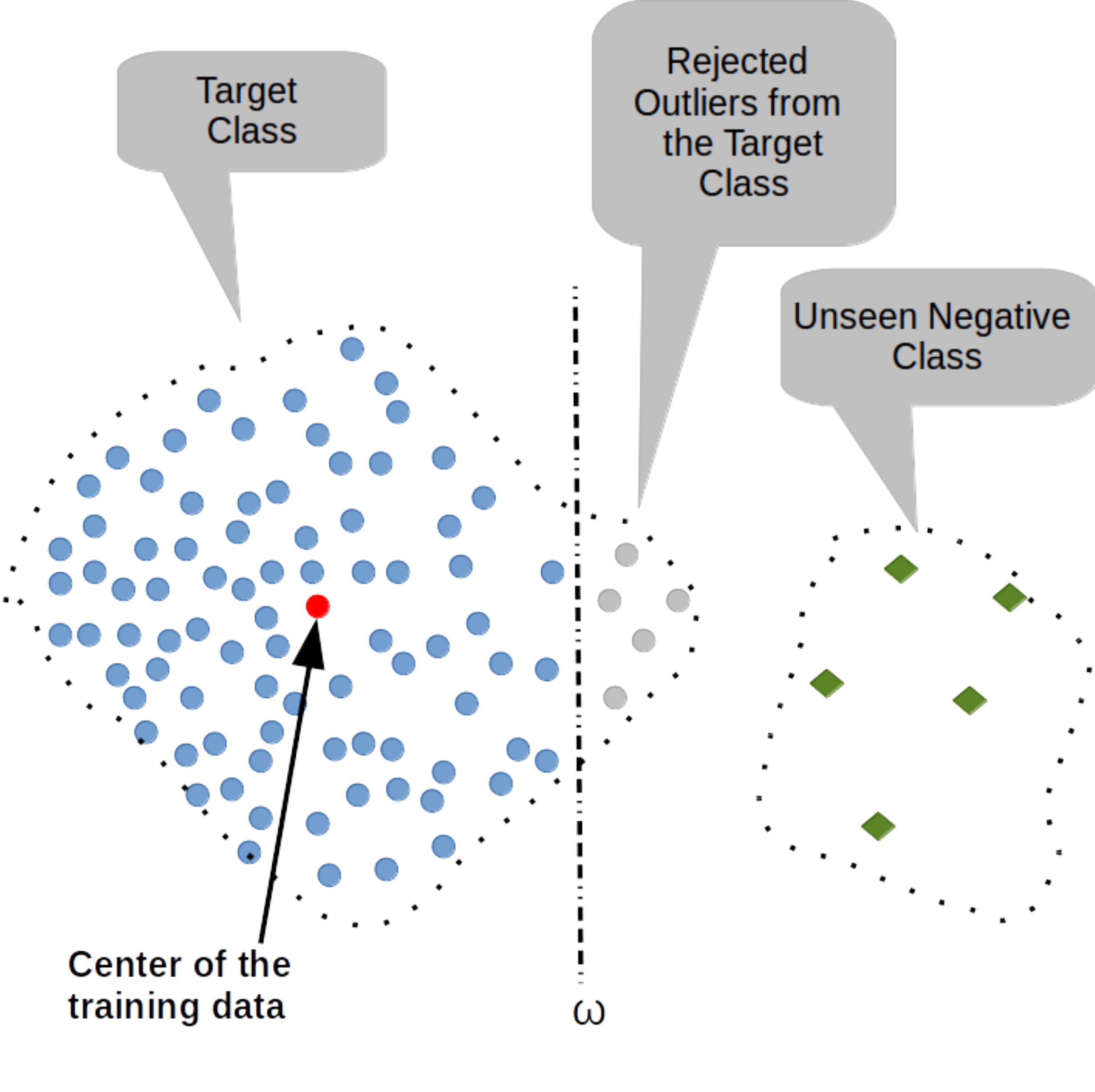}
  \caption{2-D representation of removing noisy data from the target class using IQR technique for a given $\omega$.}
  \label{fig:remove_noise}
 \end{figure}

As discussed in the previous section, noise in the target data can arise due to various reasons and its presence can adversely affect the performance of \textit{NN-d} based classifiers. 
Khan et al. \cite{khan2017detecting} show that for fall detection problem, deviant normal human motion sequences can be removed from the normal human activities using the IQR technique, which can act as a proxy for real falls and help in optimising the parameters of the probabilistic classifiers. 
In the nearest neighbour approach, we compute distances instead of probabilities and IQR technique can be adapted to work on this problem. In order to obtain distances, we use a centre based distance computation method \cite{gao2007center}. Firstly, the centre of the target data is calculated by computing the mean of all the $N$ data objects present in the target class. Then the distance of each target object is calculated from the centre, resulting in $N$ distances. These distance are then ordered and the IQR technique is applied. The quartiles of a ranked set of data values are the three points that divide the data set into four equal groups, where each group comprises of a quarter of the data. Given the distances of target data objects from the centre, the lower quartile ($Q_1$), the upper quartile ($Q_3$) and the inter-quartile range ($IQR = Q_3 - Q_1$), a point $P$ is qualified as noise of the target class, if
\begin{equation}
  P > Q_3 + \omega \times IQR \quad || \quad P < Q_1 - \omega \times IQR 
\label{eqn:iqr}
\end{equation}

where $\omega$ is the rejection rate that represents the percentage of data points that are within the non-extreme limits of distances given the distances of target training data from the centre. Based on $\omega$, the extreme values of distances that represent noisy target training data can be removed. 
Figure \ref{fig:remove_noise} shows a 2-D representation of a target class from which few noisy data are removed based on given $\omega$. The instances that are far away from the centre of the data are removed and are used as proxy for the negative class. 
The algorithmic steps to remove noisy observations from the target class is shown in the Figure \ref{algo:iqr}. These rejected noisy observation can help in the optimisation of the $J$ and $K$ neighbours or the decision threshold ($\theta$) for the OCNN using a cross-validation step, which is described next. 

\subsection{Cross Validation}
\label{sec:cv}
We now introduce a cross-validation method to optimise the nearest neighbours $J$ and $K$ for the $JKNN$ classifier when the decision threshold ($\theta$) is set to $1$. 
In a given target training data (with no negative samples), we employ the IQR technique (discussed in Section \ref{sec:iqr}) to reject some noisy observation from the target class. These rejected noisy observations serves as a proxy for the negative class. Then we employ an inner cross-validation step, with $G$ folds. We split the target and rejected noisy data such that in one fold we have target data from all the $(G-1)$ folds and test on target and rejected noisy observations of the $G^{th}$ fold. We use the one-class $JKNN$ classifier for the inner cross-validation step and test it on different values of $J$ and $K$ neighbours and the values that gives the best average value for the performance metric over all $G$ folds is chosen as the best value of the parameter. The algorithmic steps are shown in Figure \ref{algo:jknnOPT}.

\begin{algorithm}
    \DontPrintSemicolon
    \KwIn{$N$ number of $T$ target data objects with dimension $d$, rejection rate $\omega$, $G$ number of inner cross-validations}
    \KwOut{$O$: rejected noisy observations from the target class}
    Compute the centre of the target class, $\bar{T} = \sum_{i=1}^{N} \frac{T_i}{N}$\\
    Compute the distance of each target member from $\bar{T}$, $dist[i]=\Vert(\bar{T},T_i)\Vert$, where $\Vert . \Vert$ is the euclidean distance between two vectors $\bar{T}$ and $T_i$\\
    Set counter for rejected noisy targets, $counter=0$\\
    \For {every $T_i$} {
    \label{algo:stepfor}
       \If{$dist[i]>(Q3+\omega*IQR) || dist[i]<(Q1-\omega*IQR)$}{
          $O[counter]=T_i$\\
          $counter=counter+1$\\
        }
    }  
    \eIf{$counter<G$}{
      Reduce the value of $\omega$ and repeat from step \ref{algo:stepfor}
      }{
      \Return{$O$}
      }
    \caption{Removing Noisy data from the Target Class}
  \label{algo:iqr}
   \end{algorithm}

\begin{algorithm}[!ht]
\DontPrintSemicolon
\KwIn{Target training data (D), Rejected Outliers from Target ($\mathbb{N}$), Remaining Target training data {$\mathbb{D}$}, Total Number of Target training data, $N=|D|=|\mathbb{D}+\mathbb{N}|$, $J$ and $K$ nearest neighbours}
\KwOut{Optimised Neighbours, $J_{opt}$ and $K_{opt}$}
Apply inner $G$-fold cross validation, For each fold\;
{\begin{enumerate}[label=(\roman*)]
  \item Create internal training set, $\mathbb{T}$, by taking only  $\mathbb{D}$ and\\ $\mathbb{N}$ from the $(G-1)$ folds\;
  \item Create validation set, $\mathbb{V}$ by combining $\mathbb{D}$ and $\mathbb{N}$ from\\ the $G^{th}$ fold\;
  \item    \ForEach {$j \in 1,\ldots,J$}{
     \ForEach {$k \in 1,\ldots,K$}{
        \ForEach {$x \in \mathbb{V}$}{
          Apply $JKNN$ with $\theta=1$ on $\mathbb{V}$
        }
        Compute the performance metric, $gmean[j][k]\quad \forall x \in \mathbb{V}$\;
     }
   }
\end{enumerate}}
\nonl After the above step, a $gmean$ vector of length $J, K, N$ is created\;
   Find $gmean_{opt}=\mathop{\max_{i,j,n}} (gmean[j][k][n])$ \;
   Find the corresponding neighbours, i.e. $J_{opt}$ and $K_{opt}$\;
 \Return{$J_{opt}$ and $K_{opt}$}\;
\caption{Finding Optimised Nearest Neighbours for $JKNN$ with $\theta=1$}
\label{algo:jknnOPT}
\end{algorithm} 

\begin{algorithm}[!ht]
\DontPrintSemicolon
\KwIn{Target training data (D), Rejected Outliers from Target ($\mathbb{N}$), Remaining Target training data {$\mathbb{D}$}, Total Number of Target training data, $N=|D|=|\mathbb{D}+\mathbb{N}|$}
\KwOut{Empirical Threshold, $\theta_{emp}$}
Apply inner $G$-fold cross validation, For each fold\;
\label{step1}
{\begin{enumerate}[label=(\roman*)]
  \item Create internal training set, $\mathbb{T}$, by taking only  $\mathbb{D}$ and $\mathbb{N}$ from the $(G-1)$ folds\;
  \item Create validation set, $\mathbb{V}$ by combining $\mathbb{D}$ and $\mathbb{N}$ from the $G^{th}$ fold\;
  
  \item Apply $11NN$ on $\mathbb{V}$, i.e. $\forall x \in \mathbb{V}$, compute (and save) the decision threshold by computing the ratio of their first neighbours and their first neighbours (i.e., $J=K=1$) in $\mathbb{T}$\; 
\end{enumerate}}
\nonl After the above step, a decision threshold vector ($\theta_V$) of length $N$ is created\;
\ForEach {$i \in 1,\ldots,N$}{ 
 Choose $\tau=\theta_{Vi}$ as the threshold\;
\ForEach {$j \in 1,\ldots,N$}{
 Compare  $\tau$ with $\theta_{Vj}$\;
 Compute the class of each data \;
}
Compute the performance metric, $gmean_i$\;
}
Find $\theta_{emp}=\mathop{\max_{i}} (gmean_i)$ \;
 \Return{$\theta_{emp}$}\;
\caption{Finding Empirical Decision Threshold($\theta$) for $11NN$}
\label{algo:thresholding}
\end{algorithm} 


\subsection{Optimising Decision Threshold}

We discussed the relationship between $JKNN$ with $\theta=1$ and $11NN$ with variable $\theta$ in Section \ref{sec:relation}.  To optimise the decision threshold $\theta$ for $11NN$, we use the same strategy for optimising $J$ and $K$ as discussed above, that uses the rejected noise data from the positive class as a proxy for unseen negative class. Then we use a modified empirical thresholding algorithm to optimise the decision threshold $\theta$. The general idea of the original thresholding algorithm \cite{sheng2006thresholding} is to select an empirical threshold from the training instances by performing an internal cross-validation routine that optimizes the misclassification cost. 
Khan and Hoey \cite{khan2016dtfall} modified the original thresholding algorithm for a one-class decision-theoretic framework to identify unseen falls. Our method differs from both the works in two ways -- we use distance instead of the probability estimates and $gmean$ as a performance metric instead of the misclassification cost or utility (see Section \ref{sec:metric}). 

\begin{figure}[!ht]
  \centering
  \hspace{-15mm}
\includegraphics[width=10cm,height=10cm]{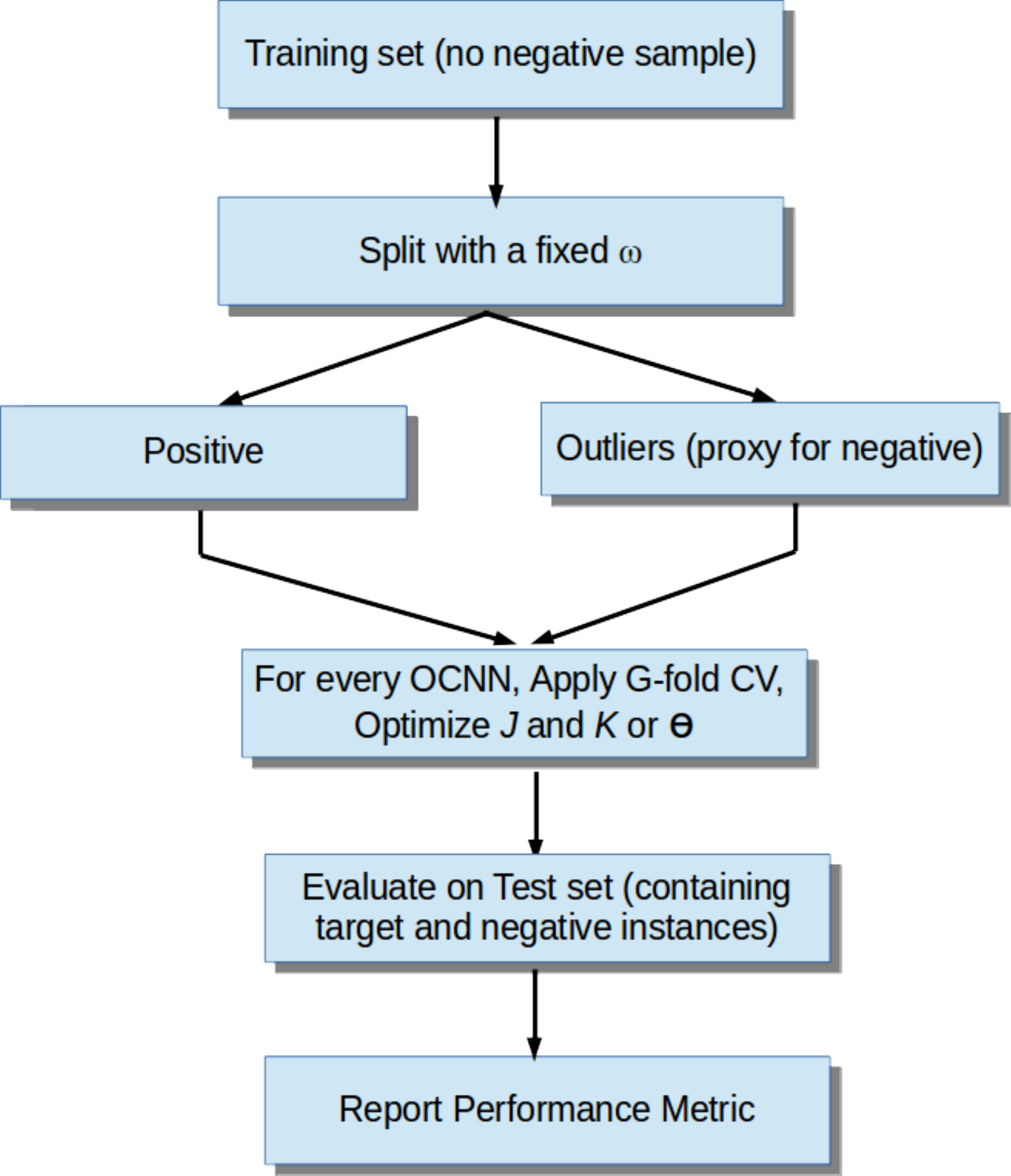}
  \caption{Block Diagram of the Procedure for Evaluating OCNN.}
  \label{fig:procedure}
\end{figure}

In the inner-cross validation step discussed in Section \ref{sec:cv}, we use a $G$ fold cross validation step. The training target data, $D$, is first split into noisy data ($\mathbb{N}$) and target data ($\mathbb{D}$, excluding noisy data) based on a given $\omega$, s.t., $|D|=|\mathbb{D}+\mathbb{N}|$. Then, an internal training set is created by taking only $\mathbb{D}$ and $\mathbb{N}$ from the $(G-1)$ folds. A validation set is created by combining $\mathbb{D}$ and $\mathbb{N}$ from the $G^{th}$ fold. 
A decision threshold is then computed for every data in the validation set. The decision threshold is defined as the ratio of distance of first nearest neighbour of a validation sample to its first nearest neighbour (i.e., $J=K=1$) in the inner training set.
After the completion of this inner cross-validation step, we will have decision thresholds corresponding to all the $\mathbb{D}$ and $\mathbb{N}$ data (that are jointly equal to the number of target instances in the training data). 
Then, we choose each of these decision thresholds, compare with all others and compute performance metrics for them. The value of decision threshold that gives the maximum value of performance metric is chosen as the empirical threshold and used with $11NN$ (instead of $\theta=1$). The algorithmic steps are presented in Figure \ref{algo:thresholding}. 
Figure \ref{fig:procedure} shows the block diagram for the overall procedure for evaluating OCNN that shows all the building blocks discussed above. 

\subsection{How many noisy observation to reject from the target class?}
As discussed earlier, noisy data can be removed from the training data using the IQR technique (Section \ref{sec:iqr}) to create a validation set for optimising the parameters of the OCNN classifier. The amount of observations to be rejected depends upon the rejection rate, $\omega$. A large value of $\omega$ results in less rejections and smaller value means more rejections. 
In the inner cross-validation method described in Section \ref{sec:cv}, the validation set should have at least one observation as a proxy for outliers in each of the inner fold $G$. 
As a rule of thumb, the value of $\omega$ should be chosen such that at least $G$ data points are rejected from the target class as proxy for outliers. Therefore, the value of $\omega$ can either be chosen using domain knowledge or hit-and-trial method. In our experiments, we mostly chose $\omega=1.5$, if that did not yield desired number of rejected targets, it is reduced until a desired number of data points from the target class are rejected to generate the validation set.

\section{Ensembles of OCNN}
\label{sec:ensemble}
 The key to build good ensembles is to create accurate and diverse classifiers \cite{Kuncheva}. Creating diverse feature spaces to create diverse classifiers is a popular method \cite{domeniconi2004nearest} for supervised KNN based ensembles. We will apply similar approaches for ensembles of OCNN classifiers, i.e. Random Subspace (RS) and Random Projection (RP).

In RS ensembles technique \cite{Ho}, in each run, a classifier is trained on a randomly selected feature subspace of the original feature space. This creates diverse classifiers that in turn creates an accurate ensemble. Random subspace technique has also be used to create ensembles for outlier detection \cite{RSONEKNN}. 
The RS method has been used to create KNN ensembles \cite{RSKNN}; however, ensembles of OCNN created by RS method have not been extensively studied. Therefore, in this paper we will explore the applicability of RS method for creating OCNN ensembles.

The RP is a method for dimensionality reduction problem \cite{randomprojectionlemma}. The RP maps a number of points in a high-dimensional space into a low dimensional space with the property that the euclidean distance of any two points is approximately preserved through the projection. 
Using matrix notation where $D_{d\times n}$ is the original set of $n$,  $d$ dimensional observations. The projection of the data onto a lower $p$-dimensional space is defined as
\begin{equation}
  D_{p\times n}^{RP} = R_{p\times d}D_{d\times n},   
\end{equation}

where $R_{p\times d}$ is a random matrix  whose rows have unit lengths and $D_{p\times n}^{RP}$ is the new ${p\times n}$ projected matrix. 
The elements $r_{ij}$ of a random matrix $R$ can be computed as \cite{RPmatrix}:

\begin{equation}
    r_{ij}= \sqrt{3}
    \begin{cases}
      +1, & \text{with probability}\ \frac{1}{6} \\
      0, & \text{with probability}\ \frac{2}{3} \\
      -1, & \text{with probability}\ \frac{1}{6} \\
    \end{cases}
  \end{equation}

Different random projections of a dataset create different new datasets with different features. 
RP has been used for creating KNN ensembles for multi-class classification  \cite{RPnsemblenearestneigh}. RP has also been used to approximate the convex-hull in high dimensional spaces for one-class problem \cite{RPONECLASSENSEm}. However, in this method RP is used to create two or three dimensional low-dimensional spaces. Hence, there may be a large loss of information. The conceptual similarity between KNN and OCNN methods motivate us to employ RP approach for creating ensembles of OCNN methods. 

There are three major steps in making RS or RP ensemble for OCNN. Firstly, apply RS or RP on the target training data $L$ times to create $L$ datasets and apply the same transformations on the test data. Secondly, for each $L$ transformed target dataset, perform inner-cross validation to find the best values of parameters, i.e., $J$, $K$ or $\theta$ (as discussed in Section \ref{sec:cv}) and predict labels for each test sample. Lastly, combine all the $L$ decisions for each test sample using majority voting to obtain final class label.
The algorithm for OCNN ensembles by RS or RP (with parameter optimisation) is illustrated in Figure \ref{algo:ensemble}. 

\begin{algorithm}
\DontPrintSemicolon
\KwIn{Target Training Data (D, positive data only), Testing data (T, both positive and negative data), the size of the ensemble is $L$,} 
 
 Apply RS or RP on D to create $L$ different datasets $D_i$ where {$i \in 1...L$} and  apply the same transformations on T to create $T_i$ datasets where {$i \in 1...L$} \;
\ForEach {$i \in 1...L$}{
      Start Inner-cross validation on the new dataset $D_i$ \;
      Find best $J, K$ or $\theta$\;
      Find predicted labels for every test sample of $T_i$ \;
}
Find final class label by using majority voting (each test sample will have $L$ labels)
\caption{OCNN ensembles with Parameter Optimisation}
\label{algo:ensemble}
\end{algorithm}

\section{Experimental Method}
\label{sec:results}

\subsection{Performance Metric}
\label{sec:metric}
Some of the popular supervised classification performance metrics, such as accuracy, precision, recall, F-measure etc., cannot be used in the OCC scenario. OCC presents a unique scenario when target data is sufficiently present during training and negative data is not. Therefore, during testing a one-class classifier, we expect to observe a highly skewed distribution of negative data w.r.t. the target data. In that sense, during testing, the problem becomes the evaluation of the performance of a classifier with severely imbalanced test set. Due to this nature of test data, conventional performance metrics (e.g. accuracy) may not be directly employed because the numbers it produce may be misleading. 
Kubat and Matwin \cite{kubat1997addressing} use the \textit{Geometric Mean} ($gmean$) of accuracies measured separately on each class i.e., it combines True Positive Rate (TPR) and True Negative Rate (TNR) (where target is the positive class and non-target is the negative class), defined as
\[ gmean = \sqrt{TPR * TNR} \] 
An important property of $gmean$ is that it is independent of the distribution of positive and negative samples in the test data. 
This measure is more useful in our application where we expect a skewed distribution of non-target w.r.t. target and we want to evaluate the performance on both the target and non-target test observations. 

\subsection{Datasets} 
\label{sec:datasets}
There are no standard datasets available for testing OCC algorithms; therefore, we test our methods on datasets that have high imbalance ratio  of the number of positive and negative labeled data. This is done to highlight the OCC problem where the data of the negative class is rare. In our experiments, only the data from the positive class is used during training and parameter optimisation i.e., no data from negative class is used during training. However, the data from negative and positive class is available during testing. We show results on $9$ benchmark datasets from KEEL repository and $6$ domain specific real-world datasets.

\subsubsection{KEEL Benchmark datasets} KEEL dataset repository \cite{alcala2011keel} provides several datasets for testing machine learning algorithms along with a separate section for imbalanced data. We choose $9$ datasets that have real valued attributes and high imbalance ratio. The details of these datasets are shown in Table \ref{tab:keel}.

\begin{table} [!ht]
\centering
  \begin{tabular}{|p{2.5cm}|p{1.7cm}|p{1.5cm}|p{1.5cm}|}\hline
    \textbf{Dataset}& \textbf{\#Attributes}& \textbf{\#Instances}& \textbf{Imbalance Ratio} \\ \hline
    yeast6                      &8        & 1484 &41.4  \\ \hline
    yeast5                      &8        & 1484 &32.73  \\ \hline
    yeast4                      &8        & 1484 &28.1  \\ \hline
    glass5                      &9        & 214  &22.78  \\ \hline
    glass4                      &9        & 214  &15.47  \\ \hline
    glass2                      &9        & 214  &11.59  \\ \hline
    winequality-red-3-vs-5 (Wine1)    &11        & 691  &68.1  \\ \hline
    winequality-white-3-9-vs-5 (Wine2) &11        & 1482 &58.28  \\ \hline
    winequality-red-8-vs-6-7 (Wine3)  &11        & 855  &46.5 \\ \hline
  \end{tabular}
  \caption{KEEL Benchmark Datasets.}
  \label{tab:keel}
\end{table}

\subsubsection{Domain-specific datasets}

  \subsubsection*{German Aerospace Centre (DLR) \cite{dlr65511}} This dataset is collected using an inertial measurement unit. 
  The dataset contains samples from 19 people of both genders of different age groups. The data is recorded in indoor and outdoor environments under semi-natural conditions. The sensor is placed on the belt either on the right or the left side of the body or in the right pocket in different orientations. The dataset contains labeled data of 
  $7$ activities: Standing, Sitting, Lying, Walking (up/downstairs, horizontal), Running/Jogging, Jumping and Falling. 
  We only use the data from accelerometer and gyroscope. One of the subjects did not perform fall activity; therefore, their data is omitted from the analysis.
  
  \subsubsection*{MobiFall (MF) \cite{mobifall}} This dataset is collected using a Samsung Galaxy S3 mobile by using the integrated 3D accelerometer and gyroscope. The mobile device was placed in a trouser pocket freely chosen by the subjects in random orientations. For falls, the subjects placed the mobile phone in the pocket on the opposite side of falling direction. All falls were monitored to be done in a specific way. 
  The mean sampling of $87$ Hz is reported for the  accelerometer and 200Hz for the gyroscope. The dataset is collected from $11$ subjects performing various normal and fall activities and $2$ subjects only performing falls activity; therefore, they are removed from the analysis. The following $8$ normal activities are recorded in this dataset: step-in car, step-out car, jogging, jumping, sitting, standing, stairs (up and down grouped together) and walking. Four different types of falls are recorded -- forward lying, front knees lying, sideward lying and back sitting chair. These data from different types of falls are joined together to make one separate class for falls.
 
 \subsubsection*{Coventry Dataset (COV) \cite{Ojetol2015data}} This dataset is collected using two SHIMMER\texttrademark sensor nodes strapped to the chest and thighs of subjects that consists of a 3D accelerometer, 3D gyroscope, and a Bluetooth device \cite{Burns2010SHIMMER}. The data was gathered at $100$ Hz. Two protocols were followed to collect data from subjects. In Protocol 1, data for four types of falls, near falls, falls induced by applying a lateral force and a set of ADL (standing, sitting, walking and lying) is collected.  Protocol 2 involved ascending and descending stairs. $42$ young healthy individuals simulated various ADL and fall scenarios, with $32$ took part in Protocol $1$ and $10$ in Protocol $2$. For Protocol $1$, the  activities were collected in a real-life circumstances. The following normal ADL were collected in Protocol $1$ -- standing, lying, sitting on a chair or bed, walking, crouching and near falls. Six types of fall scenarios are captured -- forward, backward, right, left, real fall-backward and real fall forward. The data for real fall-backward/forward is collected by standing the subject on a wobble board while they are blindfolded and try to balance themselves, then they are pushed from behind/front to fall forward/backward onto a cushion and remain lying down for 10 seconds. These data from different types of falls are joined together to make one separate class for falls. The subjects for Protocol $2$ did not record corresponding fall data; therefore, it is not used. In our analysis, we used accelerometer and gyroscope data from the sensor node strapped to the chest.


\subsubsection*{Aging-related bugs and software complexity metrics}

This dataset contains information on the aging-related bugs found in the Linux kernel and MySQL DBMS open source projects \cite{cotroneo2013predicting}. This dataset is meant to investigate defect prediction for aging-related bugs by using software complexity metric and machine learning techniques. The data repository contains several datasets and we choose two of them, namely: Linux Drivers (LD) and MySQL InnoDB (MI) . 
The reason to use this dataset is to motivate OCC applications, where the occurrence of the negative class (i.e. defect) is rare than normal functioning of the software. In the LD dataset, $31$ attributes have zero values and they are removed. In the MI dataset, $30$ attributes have zero values that are removed. One attribute has all zero values except at $2$ instances; however, during $5$-fold cross-validation, all the instances with zero attribute value may appear and will cause problem in the normalisation of the data. Therefore, this attribute is also removed.

\subsubsection*{Breast Cancer Dataset (BC)}
The original BC dataset is modified by the German Research Centre of Artificial Intelligence \cite{Markus2015} to use for unsupervised outlier detection and is highly imbalanced. 

The details of these domain-specific datasets are shown in Table \ref{tab:domain}.

\begin{table} [!ht]
\centering
  \begin{tabular}{|p{2.5cm}|p{1.7cm}|p{1.5cm}|p{1.5cm}|}\hline
    \textbf{Dataset}& \textbf{\#Attributes}& \textbf{\#Instances}& \textbf{Imbalance Ratio} \\ \hline
    DLR         & 31\cite{khan2016classification}   	& 26660 & 316.38 \\ \hline
    MF		& 31\cite{khan2016classification}	& 5918	&  11.13 \\ \hline
    COV		& 31\cite{khan2016classification}	& 13300	&  13.65 \\ \hline
    LD		& 51					& 2292	& 253.66 \\ \hline
    MI		& 51					& 402	&  11.56 \\ \hline
    BC		& 30					& 367	&  35.70 \\ \hline
  \end{tabular}
  \caption{Domain Specific Real-world Datasets.}
  \label{tab:domain}
\end{table}

\subsection{Experimental Setup}
To perform the experiment, we set the following values of the parameters
\begin{itemize}[leftmargin=*]
  \item $F$-fold cross-validation is used in the experiment to report the average performance metric.  We split the dataset such that target data from $(F-1)$ partitions are joined together and outliers from $(F-1)$ partitions are ignored s.t. this set is used during training and the remaining targets and outliers of the $F^{th}$ fold are used during testing. The value of $F$ is set to $5$ in our experiments. 
  \item The value of $G$-fold inner cross-validation for parameter optimisation is set to $2$.
  \item The minimum and maximum number of $J$ and $K$ neighbours to optimise in OCNN is set to $1$ and $10$. For the DLR, MF and COV dataset, this number is set to $5$ because we could not obtain results in reasonable time with large number of nearest neighbours due to large data size.
  \item Size of RS is set to $50\%$ and $75\%$ of the total size of attributes that are selected randomly \cite{Ho}.
  \item Size of the ensemble for random subspace, random projection is set to $25$ \cite{RLO}.
  \item To keep the almost all the information of the data in the new feature space, the dimension of new dataset created by RP is kept same as the original dataset \cite{randomprojectionlemma}.
  \item Rejection rate, $\omega$, is set to $1.5$. For some datasets, $\omega$ is gradually decreased until all the classifiers are able to reject at least $5$ instances from the positive class (corresponding to the number of inner $G$-cross-validation folds).
  \item The data is normalised using the min-max method to lie within $[0, 1]$ \cite{witten2005data}, i.e., $y_i= \frac{x_i-min(x)}{max(x)-min(x)}$, where $x=(x_1,x_2,\ldots,x_n)$, $n$ is the number of training data and $y_i$ is the normalised value.

\end{itemize}

We performed two kinds of experiments with the above mentioned datasets\footnote{The source code for evaluating different OCNN methods is available at https://github.com/titubeta/jknn}:

\begin{enumerate}[leftmargin=*]
  \item To compare the performance of different types of single OCNN methods i.e., $11NN$ vs $11NN(\theta)$ vs $JKNN$, where $11NN$ means $11NN$ at $\theta=1$, $11NN(\theta)$ means $11NN$ for optimised $\theta$ and $JKNN$ means $JKNN$ for optimised $J$ and $K$ and $\theta=1$.
  \item 
  To investigate the performance of  various types on OCNN methods with their RS and RP ensembles.
\end{enumerate}

\section{Results}
\label{sec:experiment}

\subsection{KEEL Benchmark Datasets} 

The results of the first experiment on the KEEL benchmark datasets are presented in Table \ref{OCNNensembleKEELdata} under the column `single' for each of the specific OCNN method i.e., $11NN$, $11NN$($\theta$) and $JKNN$. The results show that $11NN$ performed best ($gmean$) for $6$ out of $9$ datasets. We observe that, $11NN$ gives high values of TNR for all the datasets, whereas $11NN(\theta)$ and $JKNN$ produced better TPR in comparison to $11NN$. This means that the class boundary set by $11NN(\theta)$ and $JKNN$ favours accepting test samples as member of target class resulting in fewer false positives, whereas the boundary for $11NN$ favours rejecting more samples as negatives resulting in fewer false negatives. Since $gmean$ combines both TPR and TNR with equal weights, it gives a higher value of performance metric to $11NN$.
Alternatively, we can infer that $11NN(\theta)$ and $JKNN$ shift the decision boundary such that the these classifiers are biased for the positive class. 

The results of the second experiment on the KEEL benchmark datasets are presented in Table \ref{OCNNensembleKEELdata} under the columns `RS(50)', `RS(75)' and `RP' for each of the specific OCNN method. The results show that the RP ensembles of $11NN$($\theta$), $JKNN$ and $11NN$ always performed better ($gmean$) than their single counterparts. The performance advantage is more for $11NN(\theta)$ and $JKNN$. RP ensembles of $11NN(\theta)$ performed best for all datasets. Results suggest that RP ensembles show the TNR value equal to 1, which means that these ensembles predicted all the members of the negative class correctly. As discussed above, $11NN(\theta)$ and $JKNN$ give large TPR as they are biased for positive class. The combination of high TPR and high TNR for RP ensembles of $11NN(\theta)$ and $JKNN$ lead to their high $gmean$ values. 

For most of the cases, RS ensemble perform ($gmean$) worse than single classifier. Though RS ensembles of $11NN(\theta)$ and $JKNN$ generally have better TPR than that of single $11NN(\theta)$ and single $JKNN$ respectively, lower values of TNR for ensembles of $11NN(\theta)$ and $JKNN$ lead to lower values of $gmean$ for these ensembles. The RS method creates diverse classifiers; however, these classifiers may not necessarily be accurate, which affected their overall performance. The `single' version trained on all the features gave poor values for $gmean$. Training these classifiers with smaller amount of features (and creating an ensemble) still suffer from classification boundary bias and the ensemble did not improve the results. 

The performance of RP ensembles are best among all the ensemble methods for various OCNN approaches. It has been shown \cite{RPmachinelearning} that the performance of supervised KNN classifiers is not affected much in feature space created by RP. The less loss of information can be the reason for it. In our case, we are creating OCNN classifiers that have the same characteristics of supervised KNN classifiers. Hence, these OCNN classifiers trained on feature spaces created by RPs are accurate. Different RPs are creating diverse classifiers. The combination of these accurate and diverse classifiers are helping RP ensembles to correctly predict the negative class points. 
$11NN(\theta)$ may not work appropriately because it depends on the choice of $\theta$ chosen empirically from training data. Since the training data is limited, we can get sub-optimal choice for it. Therefore, in practice, the results of $11NN(\theta)$ may be different than $JKNN$ with $\theta$=1.
We observe that generally the optimised values of the $\theta$ for $11NN(\theta)$ were between $1$ and $2$. The optimal values of $J$ were between $1$ and $3$ whereas the optimised values of $K$ were always the maximum set value in the experiment. These values verify our discussion in Section \ref{sec:relation} that $11NN(\theta)$ with $\theta >$ 1, $1KNN$ and $JKNN$ (for smaller values of $J$) have similar decision boundaries.

\begin{sidewaystable}
\centering
\scriptsize\addtolength{\tabcolsep}{-5pt}
\caption{Results for different OCNN methods (single and ensembles) for KEEL datasets. Bold number show the best results.}
\begin{adjustbox}{max width=\textwidth}
 \begin{tabular}{|c|c|c|c|c|c|c|c|c|c|c|c|c|c|}
\hline
Dataset & & \multicolumn{4}{c|}{$11NN$ with optimised $\theta$} & \multicolumn{4}{c|}{$JKNN$} & \multicolumn{4}{c|}{$11NN$}
\\
\hline
 &measure &Single &RS(50) &RS(75) &RP &Single &RS(50) &RS(75) &RP&Single &RS(50) &RS(75) &RP
\\
\hline
Glass2&$gmean$&0.228(0.334)&0.309(0.286)&0.338(0.326)&{\bf 0.894(0.042)}& 0.328(0.318)&0.198(0.272)&0.326(0.321)&0.876(0.043)& 0.403(0.296)
&0.528(0.128)&0.400(0.276)&0.719(0.060)
\\ &TPR&0.780(0.103)&0.872(0.067)&0.816(0.046)&0.801(0.075)&0.662(0.212)
&0.897(0.051)&0.785(0.094)&0.770(0.077)&0.500(0.110)&0.550(0.073)&0.438(0.056)
&0.520(0.086)
\\ 
&TNR&0.183(0.291)&0.183(0.170)&0.250(0.276)&1.000(0.000)&0.383(0.439)&0.116(0.162)
&0.250(0.276)& 1.000(0.000)&0.433(0.397)&0.516(0.190)&0.466(0.402)& 1.000(0.000)
\\
\hline
Glass4&$gmean$&0.291(0.277)&0.374(0.360)&0.593(0.378)&{\bf 0.876(0.184)}&0.478(0.327)&0.595(0.351)&0.579(0.399)
&0.831(0.248)&0.480(0.343)&0.643(0.198)&0.625(0.187)&0.601(0.230)
\\
&TPR&0.695(0.330)&0.880(0.227)& 0.895(0.081)&0.795(0.278)&0.710(0.344)&0.830(0.298)&0.755(0.310)&0.740(0.331)
&0.420(0.240)&0.500(0.215)&0.420(0.205)&0.405(0.233)
\\
&TNR&0.333(0.408)&0.300(0.298)&0.533(0.447)&1.000(0.000)&0.533(0.380)&0.600(0.365)
&0.666(0.4714)& 1.000(0.000)&0.800(0.447)&0.900(0.223)&1.000(0.000)& 1.000(0.000)
\\
\hline
Glass5&$gmean$&0.576(0.326)&0.139(0.312)&0.137(0.308)&{\bf 0.913(0.097)}&0.590(0.116)
&0.413(0.377)&0.399(0.364)&0.821(0.244)&0.596(0.159)&0.546(0.309)&0.614(0.179)&0.626(0.184)
\\&TPR&0.842(0.149)&0.925(0.113)&0.891(0.051)&0.841(0.167)&0.688(0.308)&0.835(0.274)&0.757(0.314)
&0.722(0.324)&0.429(0.175)&0.478(0.210)&0.464(0.219)&0.420(0.198)
\\&TNR&0.500(0.353)&0.100(0.223)& 0.100(0.223)&1.000(0.000)&0.600(0.223)&0.300(0.273)
&0.300(0.273)& 1.000(0.000)&0.900(0.223)&0.700(0.447)&0.900(0.2236)& 1.000(0.000)
\\
\hline
Wine1&$gmean$&0.448(0.251)&0.350(0.196)&0.362(0.212)&{\bf 0.933(0.022)}&0.559(0.143)&0.595(0.120)
&0.629(0.057)&0.856(0.035)&0.486(0.049)&0.520(0.033)&0.459(0.048)&0.503(0.056)
\\&TPR&0.714(0.085)& 0.960(0.019)&0.854(0.038)&0.871(0.041)&0.631(0.065)&0.907(0.037)&0.711(0.031)
&0.734(0.061)&0.313(0.038)&0.324(0.046)&0.231(0.0411)&0.256(0.056)
\\
&TNR&0.360(0.219)&0.160(0.089)&0.200(0.141)&1.000(0.000)&0.520(0.228)&0.40(0.141)&0.560(0.089)
& 1.000(0.000)
&0.760(0.089)&0.840(0.089)&0.920(0.109)& 1.000(0.000)
\\
\hline
Wine2&$gmean$&0.379(0.215)&0.115(0.257)&0.489(0.295)&{\bf 0.946(0.040)}&0.503(0.291)
&0.112(0.251)&0.459(0.277)&0.860(0.047)&0.476(0.269)&0.452(0.094)
&0.416(0.237)&0.572(0.060)
\\&TPR&0.733(0.106)& 0.972(0.032)&0.856(0.082)&0.897(0.075)&0.615(0.145)
&0.911(0.040)&0.709(0.056)&0.741(0.081)&0.369(0.059)&0.372(0.036)&0.317(0.012)&0.330(0.068)
\\&TNR&0.233(0.136)&0.066(0.149)&0.350(0.252)& 1.00(0.000)&0.550(0.389)&0.066(0.1491)&0.383(0.273)
& 1.000(0.000)&0.750(0.433)&0.566(0.199)&0.683(0.4099)& 1.000(0.000)
\\
\hline
Wine3&$gmean$&0.304(0.435)&0.334(0.468)&0.456(0.429)&{\bf 0.940(0.020)}&0.407(0.389)
&0.540(0.302)&0.617(0.044)&0.876(0.036)&0.512(0.119)&0.546(0.069)&0.525(0.054)
&0.533(0.053)
\\&TPR&0.772(0.1343)& 0.970(0.016)&0.901(0.024)&0.885(0.039)&0.732(0.056)
&0.919(0.032)&0.766(0.106)&0.769(0.064)&0.333(0.050)&0.341(0.044)&0.277(0.055)
&0.286(0.055)
\\&TNR&0.300(0.447)&0.300(0.447)&0.400(0.418)& 1.000(0.000)&0.400(0.418)
&0.400(0.223)&0.500(0.000)& 1.000(0.000)&0.800(0.273)&0.900(0.223)
&1.000(0.000)& 1.00(0.000)
\\
\hline
Yeast4&$gmean$&0.546(0.107)&0.000(0.000)&0.211(0.199)&{\bf 0.983(0.006)}&0.538(0.167)
&0.147(0.205)&0.520(0.128)&0.945(0.008)&0.526(0.103)&0.287(0.173)&0.541(0.066)&0.682(0.025)
\\&TPR&0.819(0.073)& 0.993(0.008)&0.960(0.023)&0.967(0.012)&0.813(0.041)&0.983(0.013)
&0.907(0.034)&0.893(0.015)&0.463(0.028)&0.901(0.011)&0.537(0.024)&0.466(0.034)
\\&TNR&0.376(0.136)&0.000(0.000)&0.080(0.083)& 1.000(0.000)&0.378(0.182)&0.056(0.082)
&0.314(0.132)& 1.000(0.000)&0.6145(0.203)&0.118(0.084)&0.552(0.127)& 1.000(0.000)
\\
\hline
Yeast5&$gmean$&0.329(0.227)&0.164(0.228)&0.366(0.222)&{\bf 0.980(0.011)}&0.431(0.160)
&0.159(0.223)&0.430(0.176)&0.942(0.024)&0.541(0.097)&0.620(0.082)&0.536(0.108)
&0.646(0.018)
\\&TPR&0.840(0.053)&0.986(0.013)&0.952(0.024)&0.961(0.022)&0.811(0.049)&0.971(0.013)
&0.913(0.021)&0.888(0.045)&0.421(0.030)&0.808(0.038)&0.460(0.027)&0.418(0.023)
\\&TNR&0.177(0.168)&0.069(0.101)&0.183(0.129)& 1.000(0.000)&0.250(0.182)
&0.066(0.099)&0.227(0.192)& 1.000(0.000)&0.708(0.227)&0.480(0.112)&0.636(0.213)
& 1.000(0.000)
\\
\hline
Yeast6&$gmean$&0.335(0.223)&0.000(0.000)&0.000(0.000)&{\bf 0.981(0.010)}&0.361(0.238)
&0.000(0.000)&0.071(0.160)&0.940(0.023)&0.462(0.065)&0.072(0.162)&0.440(0.071)
&0.648(0.017)
\\&TPR&0.823(0.073)&0.993(0.007)&0.959(0.022)&0.964(0.019)&0.806(0.058)
&0.991(0.009)&0.901(0.027)
&0.883(0.043)
&0.429(0.038)
&0.936(0.026)&0.534(0.020)&0.421(0.022)
\\&TNR&0.200(0.2167)&0.000(0.000)&0.000(0.000)& 1.000(0.000)&0.228(0.216)
&0.000(0.000)&0.0286(0.063)&1.000(0.000)&0.514(0.162)&0.028(0.063)&0.371(0.127)
& 1.000(0.000)
\\
\hline
\end{tabular}
\end{adjustbox}
\label{OCNNensembleKEELdata}
\end{sidewaystable}

\subsection{Domain-Specific Real Datasets}
The results of the first experiment on the DLR, MF, COV, LD, MI and BC datasets are shown in Table \ref{OCNNensemblerealdata} under the column `single' for each of the specific OCNN method i.e., $11NN(\theta)$, $JKNN$ and $11NN$. 
Similar to the results on KEEL benchmark datasets, we observe that $JKNN$ and $11NN(\theta)$ are more biased to the positive class than the $11NN$, thus resulting in higher TPR at the cost of lower TNR. However, in these datasets, single $JKNN$ always performed better than single $11NN$ in terms of $gmean$. For three datasets (DLR, COV and BC) single $11NN(\theta)$ performed better than single $11NN$. 

The results of the ensembles are shown in Table \ref{OCNNensemblerealdata} under the column `RS(50)', `RS(75)' and `RP' for each of the specific OCNN method i.e., $11NN(\theta)$, $JKNN$ and $11NN$. The results about the ensembles suggest that for four datasets (MF, DLR, COV, and BC)  RP ensembles of $11NN(\theta)$ perform the best whereas for other two datasets (LD and MI) RP ensembles of $JKNN$ perform best in terms of $gmean$.  The RP ensembles for $11NN(\theta)$, $JKNN$ and $11NN$ always gave high TNR and comparable TPR in comparison to RS ensembles. Combining both high TPR and TNR results in high $gmean$ for RP ensembles in comparison to RS ensemble. 

\begin{sidewaystable}
\centering
\scriptsize\addtolength{\tabcolsep}{-5pt}
\caption{Results for different OCNN methods (single and ensembles) for real datasets. Bold number show the best results.}
\begin{adjustbox}{max width=\textwidth}
\begin{tabular}{|c|c|c|c|c|c|c|c|c|c|c|c|c|c|}
\hline
Dataset & & \multicolumn{4}{c|}{$11NN$ with optimised $\theta$} & \multicolumn{4}{c|}{$JKNN$} & \multicolumn{4}{c|}{$11NN$}
\\
\hline
 &measure &Single &RS(50) &RS(75) &RP &Single &RS(50) &RS(75) &RP&Single &RS(50) &RS(75) &RP
\\
\hline
MF&$gmean$&0.189(0.142)&0.127(0.119)&0.137(0.126)&{\bf 0.752(0.141)}& 0.331(0.031)&0.348(0.058)&0.326(0.046)&0.473(0.065)& 0.293(0.022)
&0.189(0.041)&0.266(0.033)&0.301(0.035)
\\ &TPR&0.764(0.144)&  0.863(0.184)& 0.805(0.209)&0.818(0.174)
&0.239(0.055)&0.269(0.092)&0.232(0.070)&0.228(0.062)&0.118(0.019)&0.050(0.020)&0.097(0.024)&0.091(0.020)
\\ 
&TNR&0.071(0.065)&0.036(0.038)&0.047(0.048)&0.727(0.228)&0.469(0.055)&0.463(0.040)&0.469(0.055)&1.000(0.000)
&0.735(0.085)&0.749(0.11)&0.743(0.110)& 1.000(0.000)
\\
\hline
DLR&$gmean$&0.493(0.153)&0.548(0.074)&0.560(0.135)&{\bf 0.959(0.023)}&0.629(0.018)&0.662(0.030)&0.644(0.028)&0.650(0.040)&0.467(0.023)&0.309(0.033)&0.413(0.028)&0.428(0.035)
\\&TPR&0.927(0.028)& 0.953(0.037)&0.951(0.026)&0.943(0.023)&0.418(0.048)&0.457(0.041)&0.433(0.053)&0.424(0.054)&0.222(0.028)&0.097(0.020)&0.174(0.029)&0.184(0.031)
\\ &TNR&0.285(0.145)&0.322(0.096)&0.347(0.147)&0.976(0.052)&0.952(0.076)&0.964(0.078)&0.964(0.052)& 1.000(0.000)&0.988(0.026)
&0.988(0.026)&0.988(0.026)& 1.000(0.000)
\\
\hline
COV&$gmean$&0.602(0.036)&0.642(0.039)&0.611(0.033)&{\bf 0.858(0.036)}&0.576(0.036)&0.604(0.051)&0.571(0.043)&0.619(0.035)
&0.411(0.026)&0.262(0.028)&0.348(0.021)&0.385(0.027)
\\&TPR&0.646(0.101)&0.745(0.075)&0.731(0.077)&0.747(0.059)&0.388(0.039)&0.423(0.068)
&0.376(0.046)&0.384(0.043)&0.183(0.021)&0.071(0.016)&0.128(0.015)&0.149(0.020)
\\&TNR&0.574(0.113)&0.557(0.071)&0.514(0.059)&0.987(0.015)&0.858(0.035)&0.869(0.050)&0.870(0.050)
& 1.000(0.000)&0.926(0.025)&0.970(0.018)&0.950(0.023)& 1.000(0.000)
\\
\hline
LD&$gmean$&0.000(0.000)&0.000(0.0000)&0.000(0.000)&0.723(0.405)&0.222(0.304)&0.122(0.274)&0.227(0.311)&{\bf 0.799(0.016)}
&0.174(0.239)
&0.365(0.217)&0.170(0.232)&0.585(0.019)
\\&TPR&0.796(0.052)& 0.886(0.020)&0.818(0.037)&0.816(0.045)&0.621(0.028)&0.752(0.028)&0.663(0.039)&0.639(0.025)&0.368(0.0140)&0.346(0.031)&0.350(0.023)
&0.343(0.023)
\\
&TNR&0.000(0.000)&0.000(0.000)&0.000(0.000)&0.800(0.447)&0.200(0.2739)&0.100(0.223)&0.200(0.273)& 1.000(0.000)&0.200(0.273)
&0.500(0.353)&0.200(0.273)& 1.000(0.000)
\\
\hline
MI&$gmean$&0.302(0.289)&0.268(0.262)&0.277(0.257)&0.723(0.117)&0.442(0.275)&0.434(0.263)
&0.431(0.260)&{\bf 0.879(0.094)}&0.342(0.206)&
0.374(0.212)&0.389(0.224)&0.707(0.089)
\\&TPR&0.856(0.094)& 0.891(0.041)&0.848(0.084)&0.843(0.093)&0.851(0.040)&0.883(0.028)&0.878(0.033)&0.889(0.052)&0.616(0.080)&0.635(0.054)
&0.627(0.067)&0.616(0.059)
\\&TNR&0.181(0.194)&0.147(0.175)&0.157(0.155)&0.628(0.163)&0.304(0.235)&0.276(0.201)&0.276(0.201)& 0.876(0.170)&0.247(0.173)
&0.276(0.162)&0.304(0.2037)&0.819(0.154)
\\
\hline
BC&$gmean$&0.805(0.161)&0.896(0.119)
&0.823(0.171)&{\bf 0.939(0.034)}&0.828(0.077)
&0.889(0.036)&0.854(0.036)&0.862(0.025)&0.630(0.015)&0.596(0.015)&0.596(0.018)&0.600(0.031)
\\&TPR&0.832(0.088)& 0.910(0.049)&0.865(0.081)&0.882(0.064)&0.691(0.120)&0.792(0.064)&0.731(0.061)&0.745(0.044)&0.397(0.019)&0.355(0.017)&0.355(0.021)
&0.361(0.036)
\\&TNR&0.800(0.273)&0.900(0.223)&0.800(0.273)& 1.000(0.000)&
 1.000(0.000)&1.000(0.000)&1.000(0.000)& 1.000(0.000)& 1.000(0.000)&1.000(0.000)& 1.000(0.000)& 1.000(0.000)
\\
\hline
\end{tabular}
\end{adjustbox}
\label{OCNNensemblerealdata}
\end{sidewaystable}

\subsection{Statistical Testing}
To compare the different OCNN method's single and ensemble variants, we perform the Friedman's post-hoc test with Bergmann and Hommel’s correction \cite{calvo2016scmamp,scmamp}. This test computes the p-value for each pair of algorithms corrected for multiple testing using Bergman and Hommel's correction. Tables \ref{tab:11NN:test} shows the comparison between $11NN$ and its three ensemble methods (RS(50), RS(75), RP) across all the $15$ datasets (KEEL benchmark datasets and domain-specific datasets combined together). The significant differences (p-value $< 0.05$) are shown in bold. We observe that RP ensembles are significantly different from rest of the methods; showing their superior performance (refer to to column $11NN$ in Tables \ref{OCNNensembleKEELdata} and \ref{OCNNensemblerealdata}). Similar results were obtained for $11NN (\theta)$ and $JKNN$ classifiers (See Tables \ref{tab:11NNtheta:test} and \ref{tab:JKNN:test}). This test shows that RP ensembles perform better in comparison to other ensembles and the single versions of each of the different variants of OCNN.
We also compare the performance of RP ensembles corresponding to $11NN$, $11NN(\theta)$ and $JKNN$. Tables \ref{tab:RP:test} shows that RP ensembles of $11NN (\theta)$ is significantly different from the RP ensembles of $11NN$ and $JKNN$, which are also significantly different from each other. Referring to RP columns of $11NN(\theta)$, $JKNN$ and $11NN$ in Tables \ref{OCNNensembleKEELdata} and \ref{OCNNensemblerealdata} show that the RP ensembles of $11NN(\theta)$ perform best among other RP ensembles. It also shows that RP ensembles of $JKNN$ perform better than RP ensembles of $11NN$.

\begin{table}
\centering
 \caption{Friedman post-hoc test with Bergmann and Hommel's correction for $11NN$ and their ensembles.}
\begin{tabular}{|l|llll|}
\hline
& Single & RS(50) & RS(75) & RP \\
\hline
Single & n/a & 0.867 & 0.867 & {\bf 0.032} \\
RS(50) & 0.867 & n/a & 0.867 & {\bf 0.011} \\
RS(75) & 0.867 & 0.867 & n/a & {\bf 0.003} \\
RP & {\bf 0.032} & {\bf 0.011} & {\bf 0.003} & n/a \\
\hline
\end{tabular}
\label{tab:11NN:test}
\end{table}

\begin{table}
\centering
 \caption{Friedman post-hoc test with Bergmann and Hommel's correction for $11NN$ with optimized $\theta$ and their ensembles.}
\begin{tabular}{|l|llll|}
\hline
 & Single & RS(50) & RS(75) & RP \\
\hline
Single & n/a & 0.609 & 0.609 & {\bf 0.000} \\
RS(50) & 0.609 & n/a & 0.609 & {\bf 0.000} \\
RS(75) & 0.609 & 0.609 & n/a & {\bf 0.001} \\
RP & {\bf 0.000} & {\bf 0.000} & {\bf 0.001} & n/a \\
\hline
\end{tabular}
\label{tab:11NNtheta:test}
\end{table}

\begin{table}
\centering
 \caption{Friedman post-hoc test with Bergmann and Hommel's correction for $JKNN$ and their ensembles.}
\begin{tabular}{|l|llll|}
\hline
 & Single & RS(50) & RS(75) & RP \\
\hline
Single & n/a & 1.000 & 1.000 & {\bf 0.000} \\
RS(50) & 1.000 & n/a & 1.000 & {\bf 0.000} \\
RS(75) & 1.000 & 1.000 & n/a & {\bf 0.000} \\
RP & {\bf 0.000} & {\bf 0.000} & {\bf 0.000} & n/a \\
\hline
\end{tabular}
\label{tab:JKNN:test}
\end{table}

\begin{table}
\centering
 \caption{Friedman post-hoc test with Bergmann and Hommel's correction for OCNN RP ensembles.}
\begin{tabular}{|l|lll|}
\hline
 & 11NN & 11NN($\theta$) & JKNN \\
\hline
11NN & n/a & {\bf 0.000} & {\bf 0.002} \\
11NN($\theta$) & {\bf 0.000} & n/a & {\bf 0.045} \\
JKNN & {\bf 0.002} & {\bf 0.045} & n/a \\
\hline
\end{tabular}
\label{tab:RP:test}
\end{table}

\section{Conclusions and Future Work}
\label{sec:conclusions}
In this paper, we presented a theoretical analysis to demonstrate the generalisation and relationship among different variants of OCNN. We further enhanced the discriminatory power of OCNN classifier by creating accurate ensembles of different variants of OCNN. We presented a technique to optimise the parameters for the general $JKNN$ that only uses data from the target class as the data for the negative data may be absent during the training phase. We tested the proposed methods on $15$ benchmark and domain specific datasets. 
The results show that RP ensembles of OCNN classifiers have better performance as compared to their corresponding single OCNN classifiers.
Through these experiments, we also found that finding the optimal values of $J$ and $K$ is a time consuming task in $JKNN$ approach. However, on the basis of our experiments, we can suggest that only low values of $J$ with high values of $K$ could be used as a good choice for classifying unseen samples. 
In future, we would like to use a kernel as a distance measure for OCNN and will study the effects of their ensembles. 
A major future direction is the comparison of ensemble of OCNN classifiers with other standard one-class classifiers and their ensembles.

\bibliographystyle{IEEEtran}
\bibliography{references}

\begin{IEEEbiography}[{
\includegraphics[width=1in,height=1.25in,clip,keepaspectratio]{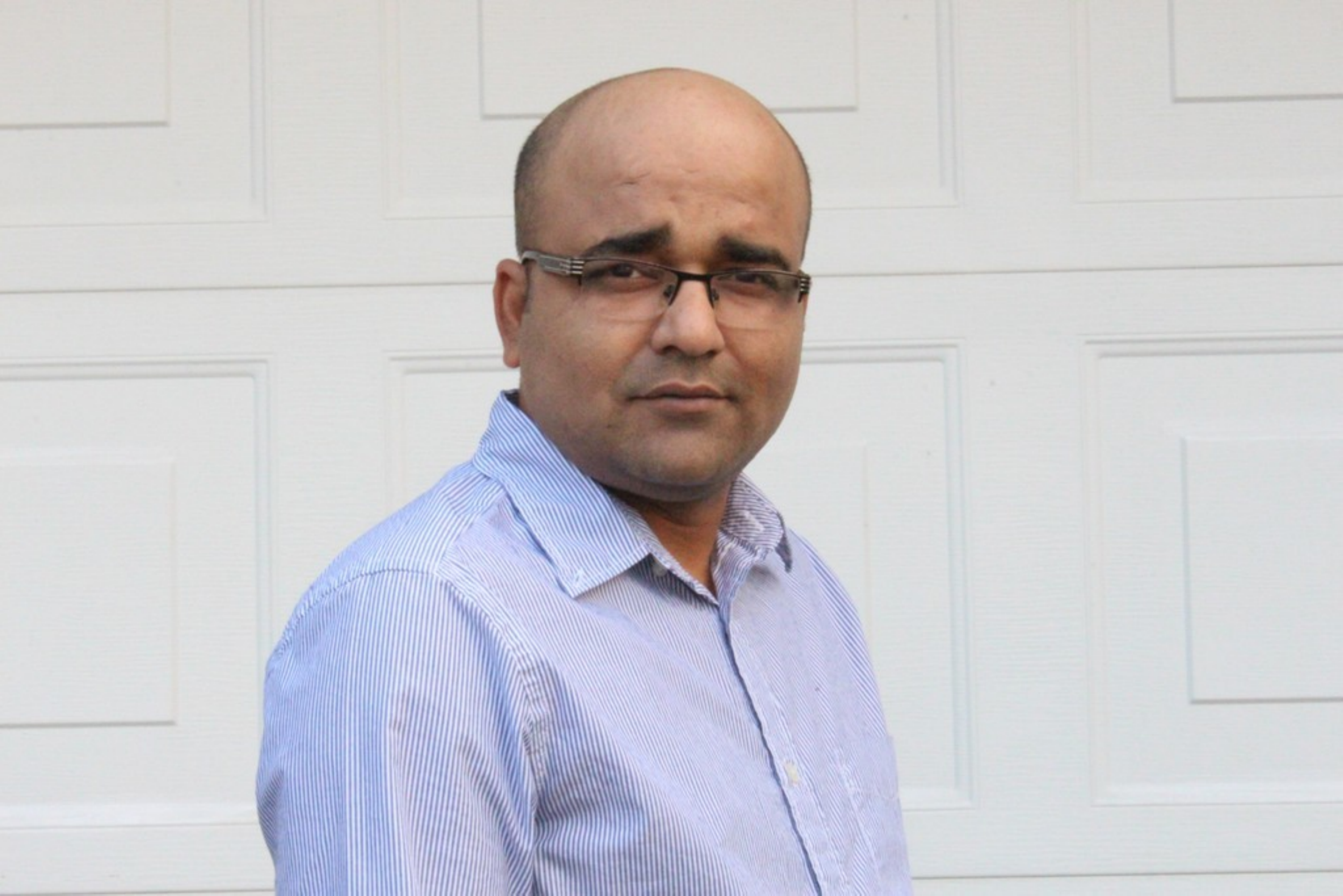}}]{Shehroz S. Khan}
Dr. Shehroz S. Khan is currently working as a Postdoctoral Researcher at the Department of Occupational Science and Occupational Therapy, University of Toronto, Canada. He earned his PhD in Computer Science with specialization in Machine Learning from the University of Waterloo, Canada. He did his Masters from National University of Ireland Galway, Republic of Ireland. 
His research interests are in machine learning, one-class classification, abnormality detection and their applications in the field of health informatics and aging. 
\end{IEEEbiography}
\begin{IEEEbiography}[{
\includegraphics[width=1in,height=1.25in,clip,keepaspectratio]{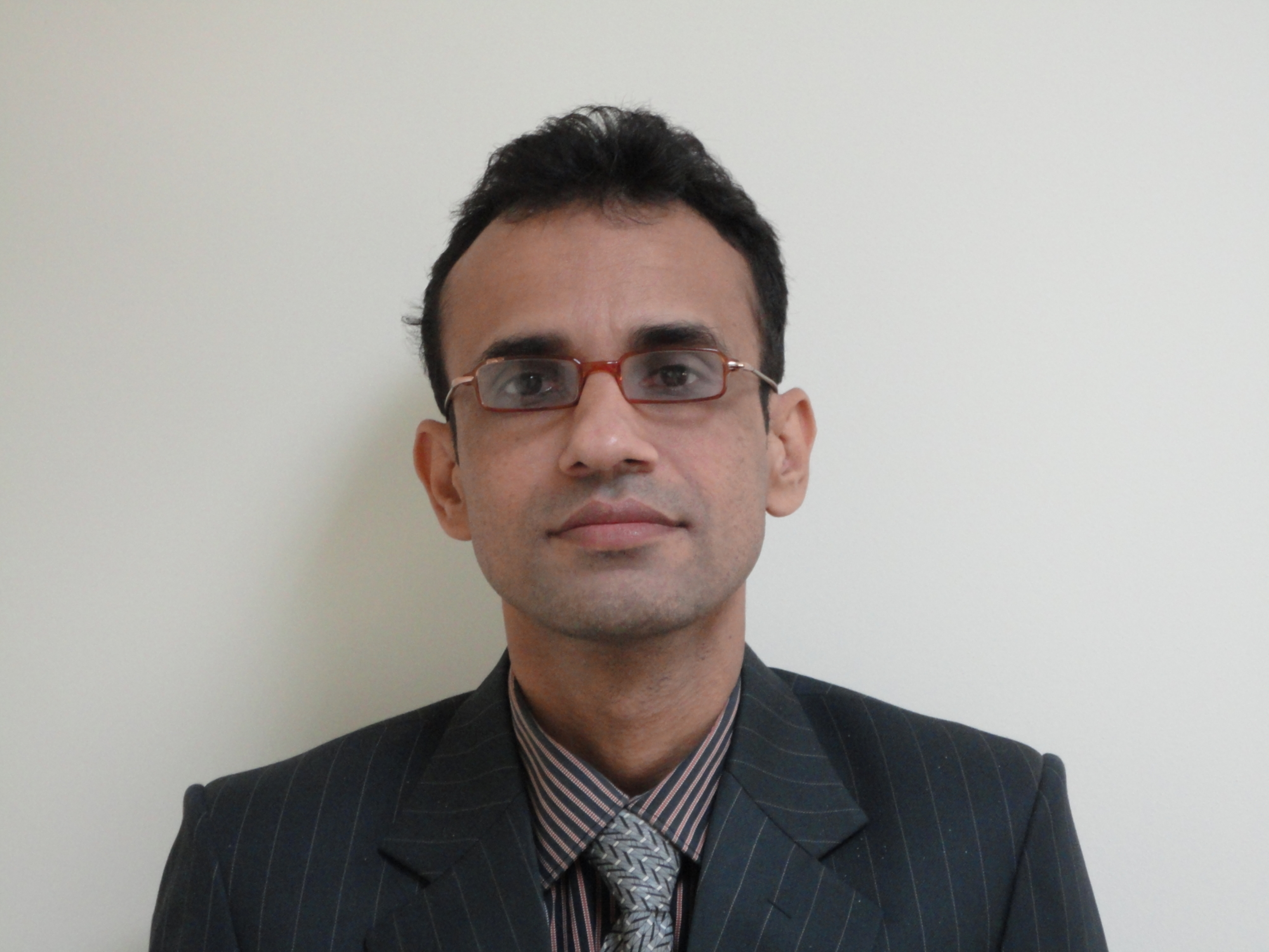}}]{Amir Ahmad}
Dr. Amir Ahmad has a PhD from University of Manchester, UK  in Computer Science. He is currently working as an Assistant Professor in the College of Information Technology, UAE University, Al Ain, UAE. His research areas are machine learning, data mining and nanotechnology.
\end{IEEEbiography}

\end{document}